\documentclass[a4paper,fleqn]{cas-dc}
\usepackage[authoryear,sort&compress]{natbib}

\usepackage{booktabs, multirow} 
\usepackage{soul}
\usepackage{changepage,threeparttable} 
\usepackage{amsfonts}
\usepackage{amsmath}
\usepackage{todonotes}
\usepackage{array}
\newcolumntype{Z}[1]{>{\centering\arraybackslash}p{#1}}

\usepackage{makecell}

\usepackage{soul}
\usepackage{color}

\usepackage{pifont}

\usepackage{url}

\newcommand{\cmark}{\ding{51}}%
\newcommand{\xmark}{\ding{55}}%

\definecolor{col_a}{HTML}{EAC2CA}
\definecolor{col_b}{HTML}{C0D8D3}
\definecolor{col_c}{HTML}{C7AFD4}
\definecolor{col_d}{HTML}{B8D9F4}
\definecolor{col_e}{HTML}{DBD9DB}
\definecolor{col_f}{HTML}{F0C594}
\definecolor{col_g}{HTML}{FEEE86}
\definecolor{col_h}{HTML}{8EC76B}

\DeclareRobustCommand{\hla}[1]{{\sethlcolor{col_a}\hl{#1}}}
\DeclareRobustCommand{\hlb}[1]{{\sethlcolor{col_b}\hl{#1}}}
\DeclareRobustCommand{\hlc}[1]{{\sethlcolor{col_c}\hl{#1}}}
\DeclareRobustCommand{\hld}[1]{{\sethlcolor{col_d}\hl{#1}}}
\DeclareRobustCommand{\hle}[1]{{\sethlcolor{col_e}\hl{#1}}}
\DeclareRobustCommand{\hlf}[1]{{\sethlcolor{col_f}\hl{#1}}}
\DeclareRobustCommand{\hlg}[1]{{\sethlcolor{col_g}\hl{#1}}}
\DeclareRobustCommand{\hlh}[1]{{\sethlcolor{col_h}\hl{#1}}}

\def\tsc#1{\csdef{#1}{\textsc{\lowercase{#1}}\xspace}}
\tsc{WGM}
\tsc{QE}

\begin{document}
\let\WriteBookmarks\relax
\def\floatpagepagefraction{1}
\def\textpagefraction{.001}

\shorttitle{Longitudinal Data and a Semantic Similarity Reward for CXR Report Generation}    

\shortauthors{Aaron Nicolson \textit{et al.}}  

\title [mode = title]{Longitudinal Data and a Semantic Similarity Reward for Chest X-ray Report Generation}  



%

\author[1]{Aaron Nicolson}[orcid=0000-0002-7163-1809]

\ead{aaron.nicolson@csiro.au}
\cormark[1]
\author[1,2]{Jason Dowling}[orcid=0000-0001-9349-2275]
\author[3,4,5]{Doug Anderson}[]
\author[1,2]{Bevan Koopman}[orcid=0000-0001-5577-3391]

\affiliation[1]{
    organization={The Australian e-Health Research Centre, CSIRO Health and Biosecurity},
    addressline={Brisbane, Australia}, 
}
\affiliation[2]{
    organization={School of Electrical Engineering \& Computer Science, University of Queensland},
    addressline={Brisbane, Australia}, 
}
\affiliation[3]{
    organization={Imaging Associates},
    addressline={Melbourne, Australia}, 
}
\affiliation[4]{
    organization={St Vincent’s Hospital},
    addressline={Melbourne, Australia}, 
}
\affiliation[5]{
    organization={Monash Health},
    addressline={Melbourne, Australia}, 
}

\begin{abstract}
Radiologists face high burnout rates, partially due to the increasing volume of Chest X-rays (CXRs) requiring interpretation and reporting. Automated CXR report generation holds promise for reducing this burden and improving patient care. While current models show potential, their diagnostic accuracy is limited. Our proposed CXR report generator integrates elements of the radiologist workflow and introduces a novel reward for reinforcement learning. Our approach leverages longitudinal data from a patient's prior CXR study and effectively handles cases where no prior study exist, thus mirroring the radiologist's workflow. In contrast, existing models typically lack this flexibility, often requiring prior studies for the model to function optimally. Our approach also incorporates all CXRs from a patient's study and distinguishes between report sections through section embeddings. Our reward for reinforcement learning leverages CXR-BERT, which forces our model to learn the clinical semantics of radiology reporting. We conduct experiments on publicly available datasets --- MIMIC-CXR and Open-i IU X-ray --- with metrics shown to more closely correlate with radiologists' assessment of reporting. Results from our study demonstrate that the proposed model generates reports that are more aligned with radiologists' reports than state-of-the-art models, such as those utilising large language models, reinforcement learning, and multi-task learning. The proposed model improves the diagnostic accuracy of CXR report generation, which could one day reduce radiologists' workload and enhance patient care. Our Hugging Face checkpoint (\url{https://huggingface.co/aehrc/cxrmate}) and code (\url{https://github.com/aehrc/cxrmate}) are publicly available.
\end{abstract}


\begin{highlights}
\item The proposed model integrates elements of the radiologist workflow and handles cases with and without prior studies.
\item A novel reinforcement learning reward using CXR-BERT is introduced to enhance clinical semantic learning.
\item Experiments on the MIMIC-CXR and Open-i IU X-ray datasets show the model generates reports more aligned with radiologists' reports than current state-of-the-art models.
\end{highlights}


\maketitle

\begin{figure}[]
    \centering
        \includegraphics[scale=1.0]{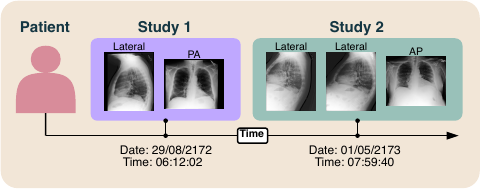}
    
    \caption{\label{fig:studies}A patient can have multiple CXR studies over time. Each study can consist of multiple images, often representing different views of the chest. Note that the year of each study has been modified for anonymisation purposes.}
\end{figure}

\begin{figure*}[]
    \centering
        \includegraphics[scale=1.25]{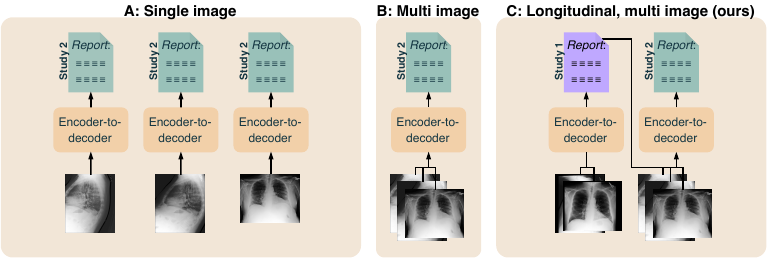}
    
    \caption{\label{fig:tasks}CXR report generation conditioned on \textsf{\textbf{A:}} a single image of a study, \textsf{\textbf{B:}} all images of a study,
and \textsf{\textbf{C:}} all images of a study, as well as the report of the previous study.}
\end{figure*}

\section{Introduction}



Burnout --- a risk factor for mental illness --- is increasingly prevalent amongst radiologists due to factors including high workloads and clinical documentation \citep{bailey_understanding_2022}. Particularly in emergency departments, it is predicted that improving radiologists' efficiency through the automation of image interpretation and radiology reporting can reduce the burden placed on them and improve patient outcomes \citep{shen_grand_2021}. With the Chest X-ray (CXR) being the most ubiquitous first-line imaging tool for chest assessment globally, automatic CXR report generation becomes a logical solution \citep{jones_chest_2021}. While current CXR report generators utilising encoder-to-decoder models are promising, a significant improvement in diagnostic accuracy is required before they can be considered for clinical trials. 

Several factors about how CXRs are interpreted in a clinical setting have been overlooked in previous approaches to CXR report generation. We outline these factors --- and how we address them to improve CXR report generation --- in Figures \ref{fig:studies} and \ref{fig:tasks}. Figure~\ref{fig:studies} shows that a CXR study for a patient can consist of multiple images. A CXR image can be a frontal view of the patient --- such as an \textit{anteroposterior} (AP) or \textit{posteroanterior} (PA) view --- or a lateral view. Radiologists will interpret all images of a study in unison before writing a report, as each view contains important information~\citep{gaber_lateral_2005}. However, this is sometimes overlooked in prior models, where a separate report is generated per image, as shown in Figure~\ref{fig:tasks} A \citep{chen_generating_2020}. This does not replicate real world radiologist reporting practice. Few models are conditioned on all images of a study if available, as shown in Figure~\ref{fig:tasks} B \citep{miura_improving_2021}. Moreover, there has been no empirical evaluation comparing multi-image to single-image CXR report generation. Hence, we provide this evaluation as a contribution of our investigation. 

Returning to Figure~\ref{fig:studies}, we note that a patient can also have multiple studies over time (e.g., Study 1 and Study 2). A radiologist utilises longitudinal data by comparing the current study with its previous, identifying key differences that can enhance diagnostic accuracy \citep{kellyChestRadiograph2012}. Although standard in the clinical setting, conditioning on longitudinal data has not been thoroughly investigated for automated CXR report generation \citep{wu_deltanet_2022}. 

\begin{figure}[]
    \centering
    
        \includegraphics[scale=0.85]{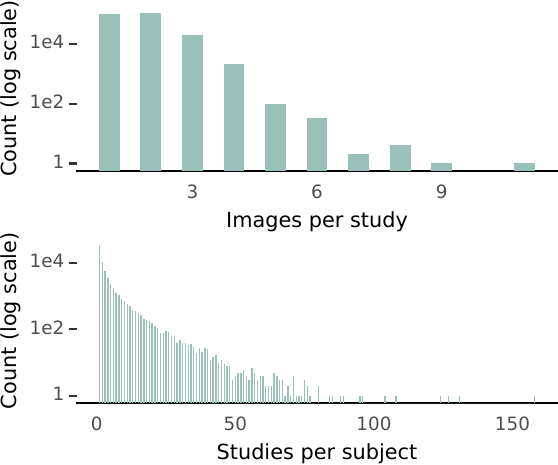}
        
    \caption{\label{fig:mimic-cxr}Histograms of the training split of MIMIC-CXR \citep{johnson_mimic-cxr-jpg_2019}. \textbf{Top}: multiple images are often taken for a single CXR study, thus motivating multi-image CXR report generation. \textbf{Bottom}: a patient often has multiple CXR studies over time, thus motivating our longitudinal, multi-image CXR report generation approach.}
    
\end{figure}

Motivated by this, we propose longitudinal, multi-image CXR report generation to improve diagnostic accuracy. Here, the model is conditioned not only on all images of a patient's current study, but also on the report from their previous study (when available), as shown in Figure~\ref{fig:tasks} C. We accomplish this in a parameter efficient manner; we adapt a multi-image CXR report generator using Low-Rank Adaptation (LoRA) \citep{hu2022lora} to accept the report of the previous study as a prompt. Evidence to support longitudinal, multi-image CXR report generation is given in Figure \ref{fig:mimic-cxr}. The top plot shows that 55\% of studies contain multiple images, while the bottom plot reveals that 50\% of patients have more than one study. This underscores that multiple images and longitudinal data can be frequently leveraged for CXR report generation. Unlike previous models that leverage longitudinal data, ours is flexible --- it can leverage longitudinal data when available. Yet, its performance is not compromised when a patient has no previous study. This is, in part, due to our training schema: our model is trained on studies where there is and is not a previous study available.

In addition to this, we investigate a new domain-specific reward for reinforcement learning. Self-Critical Sequence Training (SCST) is a reinforcement learning algorithm for image captioning that mitigates the \textit{exposure bias} problem \citep{rennie_self-critical_2017}.\footnote{Exposure bias refers to error accumulation during generation caused by the lack of exposure of a model to its own generated tokens during training \citep{rennie_self-critical_2017}.} When paired with a reward that captures the semantic similarity between the generated and radiologist reports, SCST is able to significantly improve the performance of a CXR report generator \citep{liu_clinically_2019}. The choice of reward can have a large impact on performance, as demonstrated by the state-of-the-art reward based on RadGraph \citep{delbrouck_improving_2022}. This reward makes use of named entity recognition; it compares the entities and relations between the generated and radiologist reports. 

Recently, CXR-BERT was proposed, a Transformer encoder pre-trained with contrastive representation learning that can place reports that are semantically similar close together in latent space, while placing those that are dissimilar farther apart \citep{boecking_making_2022}. We propose a reward based on CXR-BERT, to force our model to learn the clinical semantics of radiology reporting. Specifically, the reward is formed via the cosine similarity between the generated and radiologist reports in CXR-BERT's latent space. We demonstrate that this better captures the semantic similarity between the generated and radiologist reports than the entities and relations of RadGraph.




By understanding and then translating the nuances of the radiologist workflow into a succinct neural architecture, as well as introducing a reward that captures the semantics of radiology reporting, our proposed model is able to generate reports that are more aligned with radiologists' reports than current CXR report generators, such as those leveraging large language models, reinforcement learning, and multi-task learning. We evaluate our model on the publicly available MIMIC-CXR and Open-i IU X-ray datasets \citep{johnson_mimic-cxr-jpg_2019, demner-fushman_preparing_2016} with metrics shown to be more closely correlated with radiologists’ assessment of reporting \citep{Yu2022.08.30.22279318}. Furthermore, the characteristics that lend CXR-BERT to being a suitable reward for reinforcement learning also make it appropriate as a metric for CXR report generation, hence, we introduce it here additionally as a metric. Finally, there is a lack of available code and model checkpoints in the literature, making it difficult for the field to progress. To overcome this, we provide our code repository and model checkpoints in an easily accessible manner. To summarise, the main contributions of this investigation are as follows: 

\begin{enumerate}
    \item Integration of the radiologist workflow into CXR report generation; conditioning on all images of a patient's current study and the report from their previous study when available; the differentiation of report sections via section embeddings and separator tokens (described in Subsection \ref{sec:section}).
    \item A reinforcement learning reward and metric based on CXR-BERT. 
    \item A comparison between the single-image, multi-image, and longitudinal, multi-image cases; a comparison between the CXR-BERT reward and other rewards in the literature; and a comparison with state-of-the-art CXR report generators in the literature.
    \item Our Hugging Face checkpoint and code repository are publicly available:
    \begin{itemize}
        \item \url{https://huggingface.co/aehrc/cxrmate}
        \item \url{https://github.com/aehrc/cxrmate}
    \end{itemize}
    
    \item We also highlight issues pertaining to the evaluation of a large portion of CXR report generation models in the literature, in particular, where the fidelity of the labels to the original radiologist reports has been compromised due to excessive formatting. This is described in Subsection \ref{sec:section}.
    \item We develop a means of performing SCST with the generated report from the previous study, which is described in Subsection \ref{sec:gen_report_prompt}. Prompting with the \textit{generated} report from the previous study, rather than the \textit{radiologist} report from the previous study is important, as a radiologist report may not always be available in practice.
\end{enumerate}

\section{Background}
\subsection{Datasets}

The MIMIC-CXR dataset is publicly available and consists of radiographic studies performed at the Beth Israel Deaconess Medical Center in Boston, MA, between 2011--2016 \citep{johnson_mimic-cxr-jpg_2019}. Each study includes a semi-structured free-text radiology report that describes the radiological findings of the images, written by a practising radiologist contemporaneously during routine clinical care. All images and reports were de-identified to protect privacy. MIMIC-CXR is the standard dataset used for CXR report generation evaluation, due to its relatively large size and high quality \citep{Yu2022.08.30.22279318}. It is also one of the few CXR datasets that retains the metadata and radiology reports of each patients study, and is currently the sole publicly-available dataset to retain the longitudinal information between studies \citep{johnson_mimic-cxr-jpg_2019}. We use it for model training and evaluation; how we split and format MIMIC-CXR is described in Subsection \ref{sec:dataset}.

The Open-i IU X-ray dataset is also publicly available \citep{demner-fushman_preparing_2016}. It consists of radiology reports from two large hospital systems within the Indiana Network for Patient Care, along with associated images from the hospitals’ picture archiving systems. The images and reports were de-identified automatically with subsequent manual verification. Only one study per patient was included in the dataset, where outpatient studies were targeted. Even though it is relatively small, Open-i IU X-ray is often used in pair with MIMIC-CXR for CXR report generation evaluation. Though longitudinal data cannot be leveraged with Open-i IU X-ray, we use it as a test set to assess generalisability. How we format Open-i IU X-ray is described in Subsection \ref{sec:dataset}.

\subsection{Related Work} 

CXR report generation is typically performed with an encoder-to-decoder model, with many recent studies proposing architectural improvements. \citet{chen_generating_2020} proposed a “memory-driven” Transformer decoder (R2Gen), which was later developed into a “Cross-modal Memory Network” (CMN) \citep{chen_cross-modal_2021}. \citet{nicolson_improving_2022} compared different encoder and decoder architectures and pre-trained checkpoints, where it was found that the Convolutional vision Transformer (CvT) and DistilGPT2 performed best (CvT2DistilGPT2). Following CvT2DistilGPT2, we use CvT as the encoder for our model. However, we use a randomly initialised Transformer decoder with a tokeniser formed from the reports of MIMIC-CXR to establish a domain-specific vocabulary.

Others have proposed new objectives to improve the model's understanding of the relationship between CXR and report. \citet{yan_weakly_2021} proposed a Weakly-supervised Contrastive Loss (WCL) between features of the CXR and radiologist report, where negative samples that were semantically closer to the radiologist report were given more weight. \citet{najdenkoska_uncertainty-aware_2022} forced features of the CXR and radiologist report to be aligned in a latent space by formulating the report generation task as a conditional variational inference problem. For our model, we use the standard objectives for text-to-image generation associated with teacher forcing \citep{williams_learning_1989} and Self Critical Sequence Training (SCST) \citep{rennie_self-critical_2017}.

Multiple investigations demonstrate that conditioning CXR report generation on a patient's previous study improves performance \citep{wu_deltanet_2022,dalla_serra_controllable_2023,greenspan_utilizing_2023}. Longitudinal data has also been investigated for contrastive representation learning with BioViL-T, which utilises both a CXR encoder and a radiology report encoder to complete tasks such as report retrieval \citep{bannur_learning_2023}. For our investigation, we build upon prior longitudinal CXR report generation approaches and address several of their weaknesses. For example, \citet{wu_deltanet_2022} restricted their evaluation to patients with four or more studies, while \citet{greenspan_utilizing_2023} only include patients with three or more studies from the MIMIC-CXR test set. In contrast, we do not exclude patients from MIMIC-CXR based on their number of studies --- even those with only a single study are considered. This is a more difficult task for a longitudinal model as it must handle cases that do not have a previous study. Previous longitudinal models considered only a single image from a patient's study. Furthermore, lateral views were excluded in \citet{wu_deltanet_2022,dalla_serra_controllable_2023}. Our model differs by considering all CXRs for a study, regardless of view. By performing longitudinal, multi-image CXR report generation rather than longitudinal, single-image CXR report generation, our model is better aligned with the radiologist workflow. Additionally, while previous work employed extra encoders to process the previous study, our model uses a more parameter-efficient approach by adapting a multi-image CXR report generator with LoRA to accept the previous study's report as a prompt. Furthermore, previous work conditioned only on the \textit{radiologist} report from the previous study; in our investigation, we additionally consider the case of conditioning on the \textit{generated} report from the previous study --- as a radiologist report may not always be available in practice. Moreover, previous work does not consider longitudinal data with SCST. In this direction, we develop a means of performing SCST with the generated report from the previous study, as described in Subsection \ref{sec:gen_report_prompt}. Finally, the experiments from \cite{wu_deltanet_2022,dalla_serra_controllable_2023,greenspan_utilizing_2023} are not reproducible; either there is no code, or there is no model checkpoint available, making them difficult to compare to.

Reinforcement learning with SCST has offered significant improvements in the diagnostic accuracy of CXR report generation. This is especially true when the chosen reward is able to capture the semantic similarity between the generated and radiologist reports. \citet{miura_improving_2021} proposed ${\rm fact}_{\rm ENT}$ and ${\rm fact}_{\rm ENTNLI}$, two rewards that take advantage of Named-Entity Recognition (NER). Here, the number of entity matches between the generated and radiologist reports forms the basis of these rewards. \citet{delbrouck_improving_2022} proposed a reward that was able to outperform ${\rm fact}_{\rm ENT}$ and ${\rm fact}_{\rm ENTNLI}$ by leveraging RadGraph. RadGraph is a dataset of entities and relations from 500 MIMIC-CXR radiology reports that was used to train a model to jointly predict the entities and relations from reports \citep{jain2021radgraph}. The reward compares the entities and relations between the generated and radiologist reports extracted using this model (RadGraph ER). In this study, we compare these rewards, along with others, to the proposed CXR-BERT reward and show that CXR-BERT more effectively captures the semantics of radiology reporting.

\begin{figure*}[]
    \centering
    
        \includegraphics[scale=1.0]{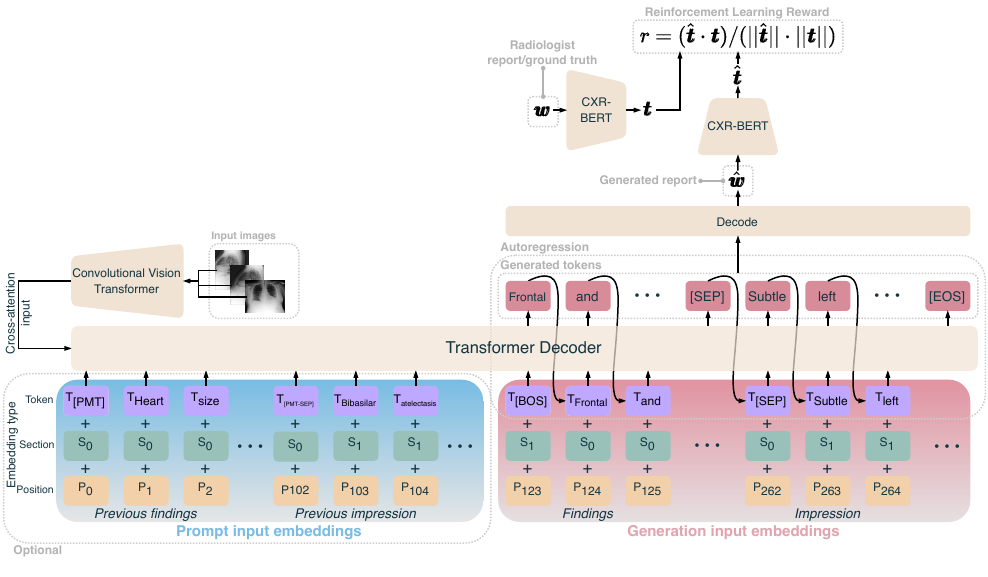}
        
    \caption{\label{fig:tokens}Our proposed model: a longitudinal, multi-image CXR report generator trained with reinforcement learning using the CXR-BERT cosine similarity reward. The findings and impression sections from the reports of the current and previous studies are differentiated by section embeddings and separator tokens. The prompt is the report of the previous study. The model is still able to generate a diagnostically accurate report even when the previous report is not available.}
        
\end{figure*}

Recently, multi-task learning has been utilised in the development of biomedical models, enhancing overall performance and generalisability by leveraging shared knowledge. With the aim of being a generalist biomedical model, Med-PaLM M is trained on multiple tasks, including classification, question answering, Visual Question Answering (VQA), report summarisation, report generation, and genomic variant calling \citep{tu_towards_2024}. With this, Med-PaLM M takes multiple modalities as input, such as images (dermatology, mammography, pathology, and chest X-rays), text (radiology reports and medical knowledge), and genomics. It has demonstrated superior performance across a variety of biomedical tasks, often outperforming specialised models and showing capabilities in zero-shot generalisation and medical reasoning. Recently, the MIMIC-CXR dataset has been utilised for multi-task learning. \citet{lee_unixgen_2023} proposed UniXGen, which leverages a Vector Quantisation Generative Adversarial Network (VQ-GAN) to discretise CXRs into discrete visual tokens. From this, UniXGen simultaneously learns to generate a report and other views of the patient. It was found that multi-task learning was beneficial to each task. LLM-CXR builds upon UniXGen by instruction-tuning a Large Language Model (LLM) conditioned on CXR features to perform either CXR report generation (CXR-to-report generation) or report-to-CXR generation \citep{lee_llm-cxr_2023}. LLaMA 2 \citep{touvron_llama_2023} was used to generate instructions from the reports of MIMIC-CXR for instruction tuning. Through this, LLM-CXR was able to outperform UniXGen on both CXR-to-report generation and VQA tasks. \citet{yang_medxchat_2023} proposed MedXChat, which jointly learned CXR-to-report generation, report-to-CXR generation, and VQA. MedXChat takes advantage of Stable Diffusion for report-to-CXR generation, and was shown to outperform LLM-CXR and UniXGen. One issue that pertains to these models is the discrepancies between how they split and format the MIMIC-CXR test set. This makes a comparison difficult. This is elaborated upon in Subsections \ref{sec:section} and \ref{sec:multi_task_results}. Nevertheless, we demonstrate that our proposed model, which utilises longitudinal data and reinforcement learning, is more diagnostically accurate than UniXGen, LLM-CXR, and MedXChat.

As of late, a lot of attention has been paid to VQA for biomedical imaging --- a more general task than CXR-based VQA. Med-Flamingo is a few-shot learner trained on image and text data from medical textbooks and PubMedCentral's OpenAccess subset \citep{moor_med-flamingo_2023}. Given a few examples, it is able to adapt to a specific medical image-text task. LLaVA-Med is also trained on PubMedCentral's OpenAccess subset \citep{li_llava-med_2023}. GPT-4 was used to generate instructions from captions to train LLaVA-Med for VQA. It was able to outperform previous state-of-the-art methods on three biomedical VQA tasks. In this study, we show that these models struggle to generalise to the task of CXR report generation, even with few-shot learning, indicating that task-specific models remain relevant.

\section{Methods} 




\subsection{Longitudinal, Multi-image CXR Report Generation}

Longitudinal, multi-image CXR report generation is defined here as conditioning the generation of a study's report on all images of a study, as well as data from previous studies. Here, we prompt the decoder with the report of the previous study when available, as shown in Figure \ref{fig:tokens}. To accommodate the prompt and the differentiation of the different sections, we adopt a schema influenced by pre-trained Transformer encoders \citep{devlin_bert_2019}. The \texttt{[PMT]}, \texttt{[BOS]}, and \texttt{[EOS]} special tokens, indicate the beginning of the prompt, the beginning of the generated report, and the end of the generated report, respectively. The \texttt{[PMT-SEP]} and \texttt{[SEP]} special tokens indicate the separation between the findings and impression sections for the prompt and generated report, respectively. Moreover, \texttt{[SEP]} allows the findings and impression sections to be extracted from the generated report. Finally, the \texttt{[NPF]} and \texttt{[NPI]} special tokens, which denote ``no previous findings''  and ``no previous impression'', respectively, are placeholders if no previous study exists. This allows the model to be trained and tested on studies that do not have a previous study. In addition, we add section embeddings to the input of the decoder to differentiate between the findings and impression sections. 

Next, we describe prompting the decoder with the \textit{radiologist} and \textit{generated} reports from the previous study. Let $\pmb w^{t} = (w^{t}_1, . . . , w^{t}_M )$ and $\hat{\pmb w}^{t} = (\hat{w}^{t}_1, . . . , \hat{w}^{t}_N )$ denote the tokens (of length $M$ and $N$) of the radiologist and generated reports for the current study $t$, respectively. Either the radiologist report from the previous study $\pmb w^{t-1}$ or the generated report from the previous study $\hat{\pmb w}^{t-1}$ can be the prompt. In real-world scenarios, conditioning on $\hat{\pmb w}^{t-1}$ might be required, as $\pmb w^{t-1}$ may not always be accessible --- such as when there is no radiologist available. During training, $\pmb w^{t-1}$ can be used as the prompt for teacher forcing or SCST, as it is readily available. However, using $\hat{\pmb w}^{t-1}$ as the prompt during training is difficult, as described in the Subsection \ref{sec:gen_report_prompt}.

\begin{table*}
    \centering
    \caption{\label{tab:cxrbert_examples}The reward between a sentence from a radiology report \textit{`Right lower lobe opacity is worrisome for consolidation, possibly due to pneumonia'} and the sentences in the table. Errors in the generated sentences are indicated by highlighting and strikethroughs.}
    \renewcommand{\arraystretch}{1.5}
    \begin{tabular}{lp{3.5in}cc}
    \toprule
        \multirow{2}{*}{Difference} & \multirow{2}{*}{Example sentences} & \multicolumn{2}{c}{Rewards}  \\ \cline{3-4}
        & & CXR-BERT & RadGraph ER\\
    \midrule
        Identical & Right lower lobe opacity is worrisome for consolidation, possibly due to pneumonia & 1.0 & 1.0 \\
        Syntactically different & Opacity in the right lower lobe is concerning for consolidation, potentially indicative of pneumonia. & 0.99 & 0.5 \\
        Incorrect location & Right \hla{middle} lobe opacity is worrisome for consolidation, possibly due to pneumonia & 0.92 & 0.83 \\
        Incorrect disease &  Right lower lobe opacity is worrisome for consolidation, possibly due to \hla{tuberculosis} & 0.86 & 0.83 \\ 
        Missing location & \st{Right lower lobe} opacity is worrisome for consolidation, possibly due to pneumonia & 0.78 & 0.5 \\
        Completely incorrect & \hla{The pulmonary venous congestive pattern persists} & -0.03 & 0.0\\
        \bottomrule
    \end{tabular}
\end{table*}

\subsection{CXR-BERT Cosine Similarity Reward} \label{sec:cxr-bert}

CXR-BERT is a Transformer encoder pre-trained in various stages on PubMed abstracts, clinical notes from MIMIC-III \citep{johnson_mimic-iii_2016}, as well as reports from MIMIC-CXR, which we denote as $E(\cdot)$ \citep{boecking_making_2022}. It has two pre-training tasks, with one being Radiology Section Matching (RSM). For RSM, the output feature vector for the \texttt{[CLS]} special token of CXR-BERT ($[E(\cdot)]_{\tt [CLS]}$) and a two-layer feedforward neural network ($P(\cdot)$) are used in series to compute features of the findings and impression sections. During RSM, a contrastive loss forces the findings and impression section features from the same report to have a higher similarity, while forcing those from different reports to have a lower similarity. 

We leverage CXR-BERT as a reward for SCST, as shown in Figure \ref{fig:tokens}. Instead of sections, we compute the similarity between the generated and radiologist reports (which include both the findings and impression sections). Features for the generated and radiologist reports are first computed as: $\hat{\pmb t} = P([E(\hat{\pmb w}^{t})]_{\tt [CLS]})$ and $\pmb t = P([E(\pmb w^t)]_{\tt [CLS]})$, respectively. Their cosine similarity then gives the reward: $r = (\hat{\pmb t} \cdot \pmb t) / (|| \hat{\pmb t}|| \cdot || \pmb t || )$. Generated and radiologist reports that are semantically similar will have a higher cosine similarity, while semantically dissimilar reports will have a lower cosine similarity. With this as the reward for SCST, the model will learn to generate reports that are semantically similar to radiologist reports.

In Table \ref{tab:cxrbert_examples} are the rewards attributed to example sentences when compared to a sentence from a radiology report. This comparison between the CXR-BERT reward and the RadGraph ER reward --- the latter being the current state-of-the-art --- aims to showcase the effectiveness of CXR-BERT in capturing the nuances of radiology reporting. Ideally, the reward should be higher when the example sentence is semantically closer to the sentence from the radiology report. The rewards given to the syntactically different example reveal that the CXR-BERT reward closely approximates the reward of the `Identical' example, showcasing its ability to capture the semantics of radiology reporting. In contrast, the RadGraph ER reward was lower than some of the examples that contained errors, suggesting that it is less effective at discerning the semantic similarities of radiology reports. Additionally, CXR-BERT is adept at identifying errors such as subtle variations in location, incorrect disease diagnoses, and missing location details. This suggests that optimising for the CXR-BERT reward can aid in learning the semantics of radiology reporting, offering an advantage over current state-of-the-art methods.

\subsection{SCST With the Generated Report From the Previous Study as the Prompt} \label{sec:gen_report_prompt}

If a model was trained with the ${\pmb w}^{t-1}$ prompt, and the $\hat{\pmb w}^{t-1}$ prompt is used at test time, the model will be susceptable to the exposure bias problem \citep{rennie_self-critical_2017}. This would be due to the model not observing the $\hat{\pmb w}^{t-1}$ prompt during training. Hence, we aim to train with the $\hat{\pmb w}^{t-1}$ prompt, thereby reducing exposure bias during testing when the $\hat{\pmb w}^{t-1}$ prompt is used. 

However, generating the $\hat{\pmb w}^{t-1}$ prompt for each example in a training mini-batch would be inefficient. To address this, we take advantage of the \textit{baseline}, which is a report generated with greedy search decoding during SCST.\footnote{During SCST, both a \textit{sample} and a \textit{baseline} are generated. The reward for the baseline is subtracted from the reward for the sample to reduce variance and stabilise training \citep{rennie_self-critical_2017}.} It can serve as the $\hat{\pmb w}^{t-1}$ prompt for the subsequent study of a patient in a later mini-batch. 

It is crucial to minimise the number of mini-batch updates between the generation of the baseline and when it is used as the $\hat{\pmb w}^{t-1}$ prompt. If too many updates occur, the baseline $\hat{\pmb w}^{t-1}$ prompt may no longer accurately represent what the model in the current mini-batch would generate, due to changes in the model parameters. To minimise this, we order a patient's studies in subsequent mini-batches (where there cannot be more than one study per patient in a mini-batch and the order of the patients for each epoch is shuffled instead of the studies). This way, the $\hat{\pmb w}^{t-1}$ prompt for an example is always from the preceding mini-batch. This approach allows the model to be trained with $\hat{\pmb w}^{t-1}$ as the prompt, which is practical since a radiologist report may not always be available for the previous study.








\begin{table*}
    \centering
    \caption{\label{tab:missing_chen_labels}An example label from \citet{chen_generating_2020}; the impression section is not included. The missing impression section from the original report is highlighted.}
    \begin{tabular}{p{3.5cm}p{6cm}p{6.5cm}}
        \toprule
        Issue & Label from \citet{chen_generating_2020} & Original findings \& impression sections \\
        \midrule
        \textbf{No impression section} (\texttt{study\_id: 59542064}). & \textbf{Findings}: the heart size appears moderately enlarged . the mediastinum demonstrates tortuosity of the thoracic aorta . there is perihilar haziness with vascular indistinctness compatible with mild pulmonary edema . hazy opacities in both lung bases likely reflect small layering bilateral pleural effusions with associated bibasilar atelectasis . no large pneumothorax is identified . there are no acute osseous abnormalities . & \textbf{Findings}: The heart size appears moderately enlarged. The mediastinum demonstrates tortuosity of the thoracic aorta. There is perihilar haziness with vascular indistinctness, compatible with mild pulmonary edema. Hazy opacities in both lung bases likely reflect small layering bilateral pleural effusions with associated bibasilar atelectasis. No large pneumothorax is identified. There are no acute osseous abnormalities. \textbf{Impression}: \hla{Mild pulmonary edema with small bilateral pleural effusions and bibasilar atelectasis.} \\
        \bottomrule
    \end{tabular}
\end{table*}

\begin{table*}
    \centering
    \caption{\label{tab:trunc_chen_labels}An example label from \citet{chen_generating_2020}. The truncation used to form the label results in information loss from the original findings section. The missing part of the findings section from the original report is highlighted.}
    \begin{tabular}{p{3.5cm}p{6cm}p{6.4cm}}
        \toprule
        Issue & Label from \citet{chen_generating_2020} & Original findings \& impression sections \\
        \midrule
        \textbf{Information loss; truncation after the 100th token} (\texttt{study\_id: 55420918}). & \textbf{Findings}: the heart is mildly enlarged with a left ventricular configuration . there is similar unfolding of the thoracic aorta . the mediastinal and hilar contours appear unchanged including a convexity along the right upper mediastinal contour . particularly since it appears stable over time it can probably be attributed to tortuosity of the great vessels . at both lung bases but more extensive on the right than left there are patchy opacities fairly streaky in nature but extensive . these are increased since the earlier examination and are accompanied by peribronchial cuffing . there is no pleural effusion or & \textbf{Findings}: The heart is mildly enlarged with a left ventricular configuration. There is similar unfolding of the thoracic aorta. The mediastinal and hilar contours appear unchanged including a convexity along the right upper mediastinal contour. Particularly since it appears stable over time, it can probably be attributed to tortuosity of the great vessels. At both lung bases, but more extensive on the right than left, there are patchy opacities, fairly streaky in nature but extensive. These are increased since the earlier examination and are accompanied by peribronchial cuffing. There is no pleural effusion or \hla{pneumothorax. Suspected mild loss in mid thoracic vertebral body heights appears unchanged and probably coincides with demineralization. The lower thoracic spine shows mild rightward convex curvature. There is wedging of an upper lumbar vertebral body which may be increased somewhat, although the apparent difference may be due to differences in orientation.} \textbf{Impression}: 1. Increasing bibasilar opacities which could be seen with lower airway inflammation or infection, although developing bronchopneumonia is not entirely excluded. 2. Mild anterior wedge compression deformity of a vertebral body at the thoracolumbar junction, likely L1; although probably chronic, potentially increased somewhat. \\
        \bottomrule
    \end{tabular}
\end{table*}

\subsection{Section Embeddings and Issues With Labels in the Literature}\label{sec:section}

Two additional factors can impact the performance of a CXR report generator, and these pertain to the labels used to develop and evaluate a CXR report generator. The first is that a radiologist's interpretation of a patients study is typically authored in multiple sections in a radiology report, including but not limited to a \textit{findings} section (which details the interpretation of a study) and an \textit{impression} section (which summarises the most important findings). However, current models are either only evaluated with labels that include only one of these sections \citep{chen_generating_2020}, or they often do not differentiate between these sections during generation \citep{thawkar_xraygpt_2023}. Only the findings section is considered in the labels of \citet{chen_generating_2020}, which are frequently used in the literature (an example is shown in Table \ref{tab:missing_chen_labels}). To account for the different sections, we introduce a separator token into the report generation process that allows the findings and impression sections of the generated report to be recovered. Moreover, we differentiate each of the sections of the report to the decoder with section embeddings --- a non-standard practice for Transformer decoders borrowed from pre-trained Transformer encoders \citep{devlin_bert_2019}. 

The second aspect is formatting that alters or removes information from the radiology reports. For example, the formatting used to form the labels of \citet{chen_generating_2020} truncates 10\% of the findings sections from the MIMIC-CXR test set (by having a maximum of 100 tokens per label). An example of the information loss that this can cause is shown in Table \ref{tab:trunc_chen_labels}; multiple important findings are lost due to the truncation. Instead, we minimise information loss by setting a maximum of 256 tokens for our labels (which includes both the findings and impression sections), which results in only 0.3\% of the reports from the test set being truncated. While this is at the cost of computational complexity, less information is lost from the original reports. Note that we only truncate the radiologist reports to 256 tokens during training; we do not truncate the radiologist reports during validation or testing (hence, the generated reports are evaluated against the full radiologist reports).

By excessively formatting the radiologist reports and not considering the impression section, the fidelity of the labels of \citet{chen_generating_2020} to the findings and impressions of the radiologists is weakened. This leads to an artificial evaluation setting which has permeated through the field of CXR report generation, as many models in the literature have been evaluated with the labels of \citet{chen_generating_2020}.

\section{Experiment Setup} 

\subsection{Dataset splitting and formatting}\label{sec:dataset}
The MIMIC-CXR dataset was used for model training and evaluation \citep{johnson_mimic-cxr-jpg_2019}, while the Open-i IU X-ray dataset was used solely for evaluation \citep{demner-fushman_preparing_2016}. The use of human data provided in these datasets was approved by the CSIRO Health and Medical Human Research Ethics Committee (2019\_086\_LR).



Currently, MIMIC-CXR is the sole publicly-available dataset to retain longitudinal information between studies. Sections from the radiologist reports of MIMIC-CXR were obtained using the official text extraction tool.\footnote{https://github.com/MIT-LCP/mimic-cxr/tree/master/txt} Studies with either a missing findings or impression section, and studies with more than five CXRs per study were removed from the official training/validation/test split. This gave a split of $57\,098$/436/280 patients, $125\,395$/991/$1\,624$ studies, and $232\,715$/$1\,837$/$2\,872$ CXRs.  Minimal formatting was applied to the radiologist reports; newline and tab characters were removed, and consecutive white spaces were replaced with a single white-space character. The order of a patient's studies was determined by the date and time provided with the metadata. The date and time for studies \texttt{57077869} and \texttt{58837588} of patient \texttt{15964158} were identical, making it impossible to determine their order. Hence, these studies, along with all of their subsequent studies were removed from the training set for the longitudinal case only, reducing the training set size to $125\,384$ studies and $232\,692$ CXRs. 

For Open-i IU X-ray, we use the entire dataset as a test set to evaluate the generalisability of the models trained on MIMIC-CXR. Findings and impression sections were extracted for each study from their corresponding XML file, where studies that did not include both a findings and impression section were excluded. No formatting was applied to either section. Longitudinal data for Open-i IU X-ray could not be leveraged as only one study per patient is available. The test set consisted of $3\,331$ studies and $6\,461$ CXRs. 

For MIMIC-CXR, we use the JPG version (MIMIC-CXR-JPG). Similarly, for Open-i IU X-ray, we use the PNG version. This is standard in the literature for CXR report generation, where the DICOM versions of both datasets are avoided. This is not ideal, as fidelity is lost. To form the JPG and PNG versions of the datasets, the authors of the respective datasets first reduced the pixel bit depth of the DICOMs from 12-16 bits to 8 bits. Furthermore, the JPG format is lossy. These factors increase the quantisation error to the DICOMs, which could be detrimental for CXR report generation. We discuss this further in Section \ref{sec:future}.



\subsection{Model} 
CvT was the encoder (specifically, CvT-21 pre-trained on ImageNet-22K and fine-tuned on ImageNet-1K at a resolution of $384\times384$) \citep{wu_cvt_2021}. Layer normalisation was applied to its last hidden state, followed by a projection to the decoder's hidden size. The encoded features for each image of a study were concatenated and fed to the cross-attention of the decoder. Each image was resized using bilinear interpolation so that its smallest side had a length of 384 and its largest side maintained the aspect ratio. Next, the resized image was cropped to a size of $\mathbb{R}^{3 \times 384 \times 384}$. The crop location was random during training and centred during testing.  Following \cite{elgendi_effectiveness_2021}, the image was rotated around its centre during training, where the angle of rotation was sampled from $\mathcal{U}{[{-5^{\circ}, 5^{\circ}}]}$. Finally, the CXR was standardised using the statistics provided with the CvT-21 checkpoint. 

For the decoder, a byte-level byte pair encoding tokeniser \citep{Wang_Cho_Gu_2020} was trained on the findings and impression sections of the training set (with a vocabulary size of $30\,000$). Tokens were fed to a randomly-initialised Transformer decoder with six layers and a language model head with a vocabulary size of $30\,000$. For the longitudinal, multi-image case, we found that training the described model on the longitudinal, multi-image CXR report generation task would not result in an improvement over the multi-image case. To ease the difficulty of learning this task, we adapt a trained multi-image CXR model to the task of longitudinal, multi-image CXR report generation using Low-Rank Adaptation (LoRA) \citep{hu2022lora}. LoRA is applied to the query and key weights of each self-attention head of the decoder with a rank of eight, an alpha of 32, and a dropout rate of 0.1. LoRA adds 147K parameters to the 80.8M parameters of the encoder-to-decoder model, where all non-LoRA parameters are frozen during fine-tuning. Greedy search and beam search with four beams were employed during validation and testing, respectively.

\subsection{Training} 
Two stages of training were performed; teacher forcing, followed by SCST. \textit{AdamW}~\citep{loshchilov_decoupled_2022} was used for mini-batch gradient descent optimisation at an initial learning rate of 5e-5 for teacher forcing and 5e-6 for SCST, with a mini-batch size of 32, for 32 epochs with teacher forcing, and for 1 epoch with SCST on $4\times$16GB NVIDIA Tesla P100 GPUs. For SCST, validation was performed every $\frac{1}{10}$ of an epoch. The validation macro-averaged CheXbert F1 was the monitored metric for checkpoint selection. For SCST, the baseline was generated with greedy search, while the sample was produced with top-\textit{k} sampling ($k=50$). During SCST, the encoder was frozen, while all parameters of the decoder were learnable (both LoRA and non-LoRA parameters). The maximum number of tokens for the generated report and the prompt was 256 each.


\begin{figure*}[]
    \centering
        \includegraphics[scale=0.825]{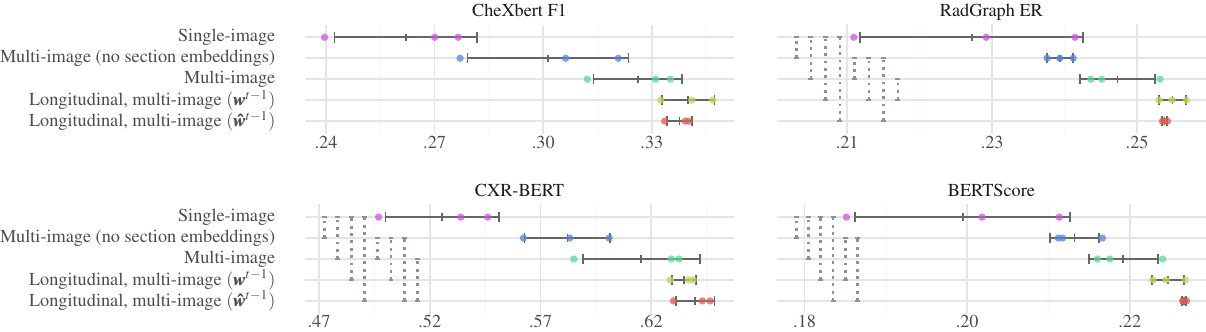}

    \caption{\label{fig:task_results}Results for the different conditioning strategies of Figure \ref{fig:tasks}. The error bars indicate the mean and standard deviation over three training runs. Dotted lines indicate a significant difference between the scores of two methods ($p<0.05$, $n=4\,872; ~1\,624~{\rm studies} \times 3~{\rm runs}$). $\pmb w^{t-1}$ indicates the radiologist report as the prompt, while $\hat{\pmb w}^{t-1}$ indicates the generated report as the prompt.}
        
\end{figure*}

\begin{figure*}[]
    \centering
    
        \includegraphics[scale=1.0, trim={0 0 0 0},clip,scale=0.85]{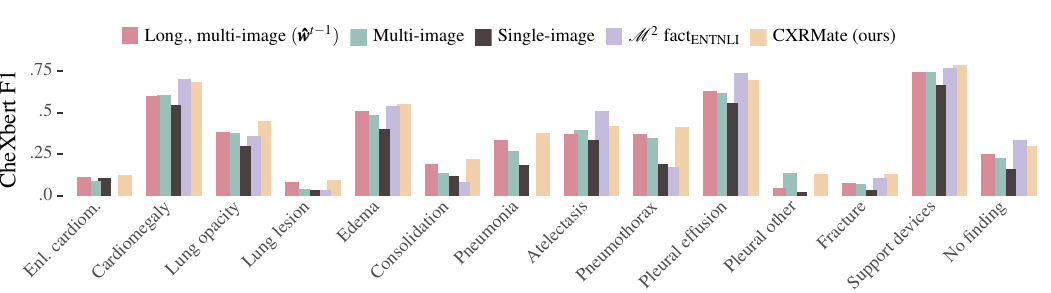}
        
    \caption{F1 for each CheXbert observation ($n=1\,624$ studies).}
    \label{fig:pathology_results}
        
\end{figure*}

\begin{figure*}[]
    \centering
    
        \includegraphics[scale=0.825]{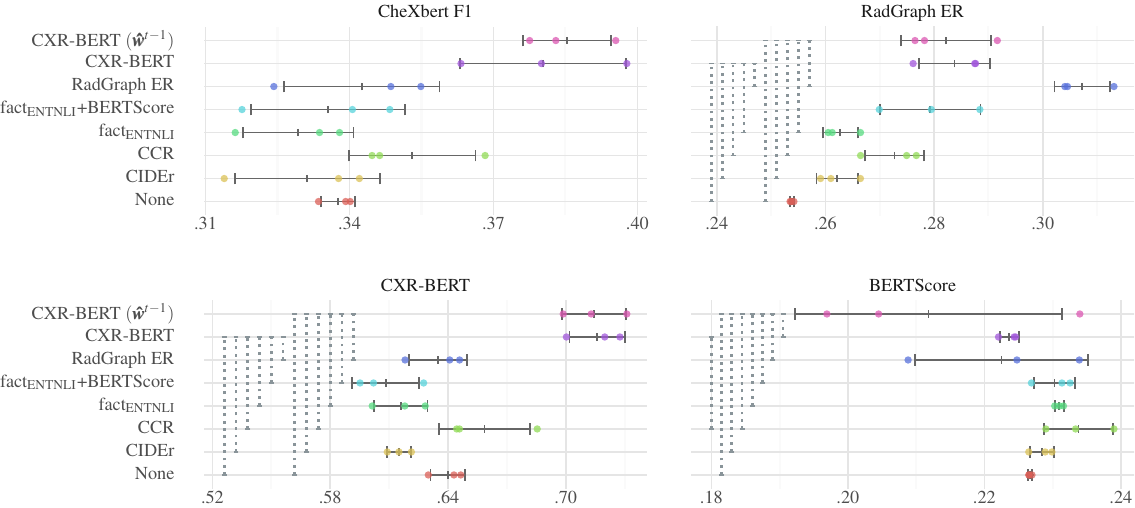}
        
    \caption{Results for each reward with the longitudinal, multi-image CXR report generator. The error bars indicate the mean and standard deviation over three training runs. Dotted lines indicate a significant difference between the scores of a method and CXR-BERT ($p<0.05$, $n=4\,872; ~1\,624~{\rm studies} \times 3~{\rm runs}$).}
    \label{fig:reward_results}
        
\end{figure*}

\begin{table*}[]
\centering
\caption{Scores of the generated findings sections (versus the findings sections from the radiologist reports) on the MIMIC-CXR test set ($n=1\,624$ studies). Each model was implemented using available code repositories and model checkpoints. }\label{tab:model_results}  

\scriptsize
\begin{tabular}{lZ{14pt}Z{29pt}Z{29pt}Z{29pt}Z{29pt}Z{29pt}Z{29pt}Z{29pt}Z{29pt}Z{29pt}}

\toprule


\multirow{3}{*}{Model} & \multirow{2}{*}{Multi-} & \multicolumn{3}{c}{CheXbert} & \multirow{2}{*}{RadGraph} & \multirow{2}{*}{CXR-} & \multirow{3}{*}{BERTScore} & \multirow{3}{*}{CIDEr} & \multirow{3}{*}{ROUGE-L} & \multirow{3}{*}{BLEU-4} \\\cmidrule{3-5}
& image &F1 &P &R & ER & BERT & & \\
\midrule



\multicolumn{11}{c}{\cellcolor[HTML]{F0F0F0}\textit{Biomedical Image VQA Models}} \\

Med-Flamingo few-shot &\xmark &0.001 &0.051 &0.000 &\cellcolor[HTML]{fef8f8}0.210 &-0.175 &\cellcolor[HTML]{ec9891}0.269 &0.042 &0.192 &0.027 \\
LLaVA-Med &\xmark &0.142 &0.202 &0.127 &0.049 &0.109 &0.000 &0.011 &0.134 &0.001 \\

\midrule
\multicolumn{11}{c}{\cellcolor[HTML]{F0F0F0}\textit{CXR report generators}} \\

R2Gen &\xmark &0.160 &\cellcolor[HTML]{f8d9d6}0.360 &0.151 &0.204 &0.377 &0.195 &0.113 &\cellcolor[HTML]{f4c5c0}0.235 &\cellcolor[HTML]{f9fdfb}0.056 \\
WCL &\xmark &\cellcolor[HTML]{d6efe3}0.208 &0.327 &\cellcolor[HTML]{d7efe3}0.199 &\cellcolor[HTML]{fef8f7}0.211 &\cellcolor[HTML]{d5eee2}0.459 &0.195 &\cellcolor[HTML]{fafdfb}0.116 &\cellcolor[HTML]{f5c7c3}0.234 &0.055 \\
CMN &\xmark &\cellcolor[HTML]{b2e0c9}0.251 &\cellcolor[HTML]{fae1de}0.353 &\cellcolor[HTML]{abddc5}0.250 &\cellcolor[HTML]{fbe9e7}0.224 &\cellcolor[HTML]{c9e9d9}0.482 &\cellcolor[HTML]{fbe7e5}0.213 &\cellcolor[HTML]{d9f0e5}0.134 &\cellcolor[HTML]{f2b9b4}0.240 &\cellcolor[HTML]{edf8f3}0.058 \\
CvT2DistilGPT2 &\xmark &\cellcolor[HTML]{acdec5}0.258 &\cellcolor[HTML]{ec9992}0.414 &\cellcolor[HTML]{acdec5}0.249 &\cellcolor[HTML]{fbe7e5}0.226 &\cellcolor[HTML]{8ed1b0}0.596 &\cellcolor[HTML]{fbe5e4}0.214 &\cellcolor[HTML]{a1d9be}0.165 &\cellcolor[HTML]{f0afaa}0.245 &\cellcolor[HTML]{c8e9d9}0.064 \\
$\mathcal{M}^2$ ${\rm fact}_{\rm ENT}$ &\cmark &\cellcolor[HTML]{bfe6d3}0.235 &\cellcolor[HTML]{f3c0bc}0.381 &\cellcolor[HTML]{b1e0c9}0.244 &\cellcolor[HTML]{f8dad8}0.237 &\cellcolor[HTML]{a4dac0}0.553 &\cellcolor[HTML]{fae2e0}0.216 &\cellcolor[HTML]{dbf1e6}0.133 &0.207 &\cellcolor[HTML]{e1f3ea}0.060 \\
$\mathcal{M}^2$ ${\rm fact}_{\rm ENTNLI}$ &\cmark &\cellcolor[HTML]{7fcba6}0.311 &\cellcolor[HTML]{ec9b95}0.412 &\cellcolor[HTML]{68c296}0.329 &\cellcolor[HTML]{e67c73}\textbf{0.320} &\cellcolor[HTML]{61c091}0.681 &\cellcolor[HTML]{e67c73}\textbf{0.289} &\cellcolor[HTML]{78c9a1}0.187 &\cellcolor[HTML]{e67c73}\textbf{0.269} &\cellcolor[HTML]{57bb8a}\textbf{0.083} \\
\textbf{CXRMate (ours)} &\textbf{\cmark} &\cellcolor[HTML]{57bb8a}\textbf{0.357} &\cellcolor[HTML]{e67c73}\textbf{0.438} &\cellcolor[HTML]{57bb8a}\textbf{0.349} &\cellcolor[HTML]{f1b3ae}0.272 &\cellcolor[HTML]{57bb8a}\textbf{0.700} &\cellcolor[HTML]{e98980}0.280 &\cellcolor[HTML]{57bb8a}\textbf{0.205} &\cellcolor[HTML]{e98b82}0.262 &\cellcolor[HTML]{6bc398}0.079 \\

\bottomrule
\end{tabular}

\end{table*}

\subsection{Comparison Models and Rewards} 
We compared the CXR-BERT reward to other rewards, including  CIDEr \citep{vedantam_cider_2015}, Clinical Correctness Reward (CCR) (with CheXbert instead of the CheXpert labeller \citep{irvin_chexpert_2019}) \citep{liu_clinically_2019}, ${\rm fact}_{\rm ENTNLI}$ and ${\rm fact}_{\rm ENTNLI}$ + BERTScore \citep{miura_improving_2021}, as well as RadGraph ER \citep{delbrouck_improving_2022}. Moreover, we compared our CXR report generator to others in the literature that had available code and model checkpoints. These included R2Gen \citep{chen_generating_2020}, $\mathcal{M}^2$ ${\rm fact}_{\rm ENT}$ and $\mathcal{M}^2$ ${\rm fact}_{\rm ENTNLI}$ \citep{miura_improving_2021}, WCL \citep{yan_weakly_2021}, CMN \citep{chen_cross-modal_2021}, and CvT2DistilGPT2 \citep{nicolson_improving_2022}. We also compared to Med-Flamingo with the following prompt for few-shot learning \citep{moor_med-flamingo_2023}:

\noindent\texttt{<image>}$F_{1}$\texttt{<|endofchunk|><image>}$F_{2}$\texttt{<|endofchunk|>}\\\texttt{<image>}$F_3$\texttt{<|endofchunk|><image>}$F_4$\texttt{<|endofchunk|>\\<image>}$F_5$ 
\texttt{<|endofchunk|><image>}. Five random CXRs were selected from the training set for the prompt, along with their corresponding findings sections ($F_1$ to $F_5$). For LLaVA-Med, `Describe the following image in detail.' was used to instruct LLaVA-Med, following \citet[Table 8]{li_llava-med_2023}. Scores for UniXGen \citep{lee_unixgen_2023}, XrayGPT \citep{thawkar_xraygpt_2023}, LLM-CXR \citep{lee_llm-cxr_2023}, and MedXChat \citep{yang_medxchat_2023} were quoted from \citet{yang_medxchat_2023}, as either their code or model checkpoint was unavailable. Each model checkpoint was trained on MIMIC-CXR, and not Open-i IU X-ray.



\subsection{Metrics} 

CheXbert \citep{smit_combining_2020}, RadGraph ER, BLEU \citep{papineni_bleu_2001}, and BERTScore F1\footnote{\texttt{roberta-large\_L17\_no-idf\_v=0.3.12(hf=4.25.1)-rescaled}} \citep{zhang_bertscore_2019} have been found to correlate with radiologists’ assessment of reporting \citep{Yu2022.08.30.22279318} and were a part of our evaluation. Additionally, we include CIDEr \citep{vedantam_cider_2015} and ROUGE-L \citep{lin_automatic_2003}, and propose to use the CXR-BERT cosine similarity as a metric. CheXbert, RadGraph ER, and CXR-BERT were intended to capture the clinical semantic similarity between the generated and radiologist reports, while BERTscore was intended to capture general semantic similarity. Finally, CIDEr, ROUGE-L, and BLEU-4 were intended to capture the syntactic similarity between the generated and radiologist reports. For the single-image models, we average the scores over all reports for a study. Following this, the average was taken over the scores for all studies for single- and multi-image models.


For CheXbert, the macro-averaged F1, Precision (P), and Recall (R) were computed between the 14 CheXbert observations extracted from the generated and radiologist reports. ``No mention'', ``negative'', and ``uncertain'' were considered negative, while ``positive'' was considered positive. Here, the true positives, false positives, and false negatives were averaged over the reports of each study for the single-image case. We also perform statistical testing; first, a Levene’s test revealed that the variances across models were not homogeneous. Next, a one-way Welch’s ANOVA determined that there was a significant difference between models. Finally, Games-Howell post hoc tests were used for pairwise testing. Statistical testing was not performed for CheXbert, as it is a binary classification-based metric.

\section{Results \& Discussion} 

\subsection{Single, Multi, \& Longitudinal Conditioning}

First, we discuss the results for single-image, multi-image, and longitudinal, multi-image CXR report generation (where each was trained with teacher forcing). The multi-image case significantly outperformed the single-image case on all metrics in Figure \ref{fig:task_results} and achieved a higher F1 on 13 of the 14 CheXbert observations in Figure \ref{fig:pathology_results}. This indicates that conditioning on all CXRs of a study (which may contain both frontal and lateral views) improves CXR report generation. This aligns with the radiologist workflow, as certain abnormalities are more easily identifiable with both views \citep{gaber_lateral_2005}.

The longitudinal, multi-image model outperformed the single- and multi-image models on all metrics in Figure \ref{fig:task_results} and on 10 of the 14 CheXbert observations in Figure \ref{fig:pathology_results}. Moreover, the difference between prompting with $\pmb w^{t-1}$ and $\hat{\pmb w}^{t-1}$ was not statistically significant, indicating that conditioning on a generated report from the previous study rather than the radiologist's report from the previous study does not hinder generation. The results indicate that conditioning on longitudinal data improves CXR report generation and diagnostic accuracy. This aligns with the radiologist workflow, as diagnostic accuracy can improve when the previous study is available for comparison \citep{kellyChestRadiograph2012}. 

The longitudinal, multi-image model was trained on studies that did and did not have a previous study. To demonstrate its flexibility, we present its performance on both scenarios in Table \ref{tab:history_impact}. For this, the model was evaluated only on studies that had a previous study available. When the longitudinal, multi-image model was prompted with the previous report, its performance increased considerably compared to when it was not prompted with the previous report. Yet, even when it was not prompted with the previous report, its performance was not compromised when compared to the multi-image model. In fact, it achieved a higher CheXbert F1, CXR-BERT score, and BERTScore. This may be due to the additional parameters associated with LoRA or because prompting with the previous report helps the model better learn the task.

\begin{table*}
    \centering
    \caption{\label{tab:history_impact}Impact of the previous study --- the longitudinal case tested with and without the previous report. Each model was tested on only the studies in the MIMIC-CXR test set that had a previous study with a findings and impression section ($n=886$ studies).}
    \begin{tabular}{lccccc}
    \toprule
        Model & Previous report (${\pmb w}^{t-1}$) & CheXbert F1 & RadGraph ER & CXR-BERT & BERTScore \\
    \midrule
        Multi-image & - & 0.329 & 0.251 & 0.609 & 0.226 \\
        Longitudinal, multi-image & \xmark & 0.315 & 0.255 & 0.636 & \textbf{0.228} \\
        Longitudinal, multi-image & \cmark & \textbf{0.345} & \textbf{0.264} & \textbf{0.650} & \textbf{0.228}\\
        \bottomrule
    \end{tabular}    
\end{table*}

Figure \ref{fig:task_results} also shows that section embeddings improve the scores on each metric (when looking at the multi-image case, with and without section embeddings).\footnote{Each model in Figure \ref{fig:task_results} uses section embeddings, except `Multi-image (no section embeddings)'.} There was little difference between the distribution of each section's vocabulary, with a Kullback--Leibler divergence of 0.04 between the findings and impression section token distributions on the training set. Yet, the findings sections are 3.4 times longer on average than the impression sections. Such a difference in length is indicative of the purpose of the impression section: to summarise the findings. We hypothesise that signalling to the decoder which section the next token belongs to via section embeddings may allow it to better understand which task it must perform: interpretation or summarisation. 

\begin{table*}[]
    \centering
    \footnotesize
    \caption{\label{tab:multi_task}Differences between the test sets and the labels of each model. Details are sourced from the respective paper of each model.}
    \begin{tabular}{lp{5cm}p{2cm}p{4cm}c}
        \toprule
        Model & Test set & Sections & Formatting  & Multi-image\\
        \midrule
        UniXGen & $4\,444$ images and $2\,733$ studies from the MIMIC-CXR test set. & Findings and impression. & Lowercase. & \cmark \\
        \midrule
        XrayGPT & MIMIC-CXR test set, exclusion schema unknown. & Combined the findings and impression sections. & Unknown. & \xmark \\
        \midrule
        LLM-CXR & $3\,530$ images; lateral views were excluded from the MIMIC-CXR test set. & Impression. & Unknown. & \xmark \\
        \midrule
        MedXChat & $3\,858$ images; studies without a findings section were excluded from the MIMIC-CXR test set. & Findings and impression. & Followed the formatting of \citet{chen_generating_2020}, differing by applying it also to the impression section. & \xmark \\
        \midrule 
        \textbf{CXRMate (ours)} & $2\,872$ images, $1\,624$ studies, and 280 patients; studies without a findings or impression section, or more than five images were excluded from the MIMIC-CXR test set. & Findings and impression. & Newline characters, tab characters, and consecutive white spaces were replaced with a single white-space character. & \cmark\\
        \bottomrule
    \end{tabular}
\end{table*}

\begin{table*}[]
\centering
\caption{CheXpert F1 scores on the MIMIC-CXR test set for the generated reports. We use CheXbert to estimate the CheXpert observations for our model \citep{irvin_chexpert_2019}. $\dagger$ indicates results quoted from \citet{yang_medxchat_2023}. Following \citet{yang_medxchat_2023}, the logical disjunction (OR) was taken between the consolidation and pneumonia observations. F and I indicate the findings and impression sections, respectively.}\label{tab:medxchat_ce_results}

\begin{tabular}{lcccccc}\toprule
&UniXGen-256$^{\dagger}$ &XrayGPT$^{\dagger}$ &LLM-CXR$^{\dagger}$ &MedXChat$^{\dagger}$ &\textbf{CXRMate (ours)} \\\midrule
Sections & F \& I & F + I & I & F \& I & F \& I \\
Samples & $2\,733$ studies & Unknown & $3\,530$ images & $3\,858$ images & $1\,624$ studies \\
Multi-image & \cmark & \xmark & \xmark & \xmark & \cmark \\
Micro &0.281 &\cellcolor[HTML]{e9f6f0}0.314 &\cellcolor[HTML]{a5dbc1}0.414 &\cellcolor[HTML]{a1d9be}0.420 &\cellcolor[HTML]{57bb8a}\textbf{0.529} \\
Macro &0.187 &\cellcolor[HTML]{fae4e2}0.227 &\cellcolor[HTML]{f3beb9}0.283 &\cellcolor[HTML]{f2b7b2}0.292 &\cellcolor[HTML]{e67c73}\textbf{0.378} \\
No Findings &\cellcolor[HTML]{b1e0c9}0.411 &\cellcolor[HTML]{c9eada}0.371 &\cellcolor[HTML]{57bb8a}\textbf{0.562} &\cellcolor[HTML]{e9f6f0}0.318 &0.280 \\
Pneumothorax &\cellcolor[HTML]{fdf2f1}0.083 &0.049 &\cellcolor[HTML]{fdf2f1}0.083 &\cellcolor[HTML]{fcefee}0.092 &\cellcolor[HTML]{e67c73}\textbf{0.388} \\
Edema &0.226 &\cellcolor[HTML]{c8e9d9}0.333 &\cellcolor[HTML]{b5e1cc}0.370 &\cellcolor[HTML]{a7dcc2}0.398 &\cellcolor[HTML]{57bb8a}\textbf{0.552} \\
Effusion &0.215 &\cellcolor[HTML]{f6cecb}0.404 &\cellcolor[HTML]{f4c1bd}0.455 &\cellcolor[HTML]{e67c73}\textbf{0.718} &\cellcolor[HTML]{e98981}0.671 \\
Consolidation OR Pneumonia &0.132 &\cellcolor[HTML]{f8fcfa}0.143 &\cellcolor[HTML]{d3ede0}0.198 &\cellcolor[HTML]{e1f3ea}0.177 &\cellcolor[HTML]{57bb8a}\textbf{0.380} \\
Lung lesion &\cellcolor[HTML]{efa7a1}0.055 &\cellcolor[HTML]{ed9d96}0.058 &0.030 &\cellcolor[HTML]{f3bcb8}0.049 &\cellcolor[HTML]{e67c73}\textbf{0.067} \\
\bottomrule
\end{tabular}
\end{table*}

\begin{table*}[]
\centering
\caption{Natural language generation metric scores on the MIMIC-CXR test set for the generated reports. $\dagger$ indicates results quoted from \citet{yang_medxchat_2023}. F and I indicate the findings and impression sections, respectively.}\label{tab:medxchat_nlg_results}
\begin{tabular}{lccccccc}\toprule
Model & Samples & Multi-image & Sections & BLEU-4 & ROUGE-L & METEOR & CIDEr \\\midrule
LLM-CXR$^{\dagger}$ & $3\,530$ images & \xmark & I &0.033 &0.245 &0.081 &\textbf{0.445} \\
\midrule
MedXChat$^{\dagger}$ & $3\,858$ images & \xmark & F \& I &\textbf{0.111} &0.264 &0.135 &0.175 \\
\midrule
UniXGen-256$^{\dagger}$ & $2\,733$ studies & \cmark & F \& I &0.101 &\textbf{0.294} &0.156 &0.138 \\
CXRMate (ours) & $1\,624$ studies & \cmark & F \& I &0.074 &0.255 &\textbf{0.158} &0.172 \\
\bottomrule
\end{tabular}

\end{table*}

\subsection{Rewards for Reinforcement Learning}

The results for each reward used with SCST --- a reinforcement learning algorithm --- are given in Figure \ref{fig:reward_results}. Each reward was evaluated using the longitudinal, multi-image model prompted with $\pmb w^{t-1}$ during training and testing, except for CXR-BERT ($\hat{\pmb w}^{t-1}$) (which was prompted with $\hat{\pmb w}^{t-1}$ during training and testing). Here, we can observe how employing SCST and optimising for an appropriate reward impacts performance. Considerable performance gains can be attained with SCST, as shown by the substantial increase in the CheXbert F1-score from `None' to `CXR-BERT' (where `None' is the model before SCST). 

As expected, CXR-BERT and RadGraph ER each performed best on the corresponding metrics that they were optimised on (CXR-BERT and RadGraph ER, respectively). Yet, CXR-BERT attained the highest CheXbert F1, indicating that rewarding based on similar latent alignments with CXR-BERT is a promising alternative to rewarding based on matching entities and relations with RadGraph ER. This also indicates that CXR-BERT is better able to capture the semantics of radiology reporting than RadGraph ER, as suggested by the aformentioned analysis concerning Table \ref{tab:cxrbert_examples}. Comparing CXR-BERT (which was prompted with ${\pmb w}^{t-1}$) to CXR-BERT ($\hat{\pmb w}^{t-1}$), their performance was similar, indicating that any errors in $\hat{\pmb w}^{t-1}$ had no significant impact on performance. We observed that the CXR-BERT reward moderately increased repetitions in the generated reports for some training runs, which likely contributed to their lower BERTScores. 

\subsection{Comparison to Other Models}

In Table \ref{tab:model_results}, the longitudinal, multi-image model trained with the CXR-BERT reward (prompted with $\hat{\pmb w}^{t-1}$ during training and testing), which we name CXRMate, was compared to other models in the literature. Note that while CXRMate generates both the findings and impression sections, the remaining models in Table \ref{tab:model_results} generate only the findings section. Therefore, only the findings section for CXRMate was evaluated in Table \ref{tab:model_results} (against the findings section from the radiologist reports), and the impression section was ignored.

CXRMate produced the highest CheXbert (F1, P, and R), CXR-BERT, and CIDEr scores. $\mathcal{M}^2$ ${\rm fact}_{\rm ENTNLI}$ is conditioned on all CXRs of a study, and was trained with a composite reward; it represents a strong benchmark. The superior performance of $\mathcal{M}^2$ ${\rm fact}_{\rm ENTNLI}$ on RadGraph ER and BERTScore is likely due to it being optimised with its ${\rm fact}_{\rm ENTNLI}$ and BERTScore reward, where ${\rm fact}_{\rm ENTNLI}$ is an NER-based reward that is similar to RadGraph ER. CXRMate outperformed $\mathcal{M}^2$ ${\rm fact}_{\rm ENTNLI}$ on 10 out of the 14 CheXbert observations in Figure \ref{fig:pathology_results}. While $\mathcal{M}^2$ ${\rm fact}_{\rm ENTNLI}$ performed well for \textit{cardiomegaly}, \textit{atelectasis}, and \textit{pleural effusion}, it failed to correctly detect \textit{enlarged cardiomediastinum} or \textit{pleural other} throughout the entire test set. CXRMate also performed best for lung lesion, a difficult and important abnormality to detect, made more difficult by the fact that it is underrepresented in the MIMIC-CXR training set \cite[Table 2]{johnson_mimic-cxr-jpg_2019}. These results indicate that CXRMate was able to generate reports that are quantitatively more aligned with those of radiologists than previous models (in terms of the CheXbert, CXR-BERT, and CIDEr scores).

The low performance of R2Gen, WCL, CMN, and CvT2DistilGPT2 could be attributed to being conditioned on a single-image, not leveraging longitudinal data, not being trained with SCST, and being trained on the truncated labels of \citet{chen_generating_2020} (described in Table \ref{tab:trunc_chen_labels}). Moreover, the results indicate that the biomedical image VQA models, namely Med-Flamingo and LLaVA-Med, struggled to interpret the CXRs. Furthermore, Med-Flamingo struggled to adapt to the task, even though it was prompted with five examples for few-shot learning.

\begin{table*}[]
\centering
\caption{Scores of the generated findings sections (versus the findings sections from the radiologist reports) on the Open-i IU X-ray dataset ($n=3\,331$ studies). Each model was implemented using available code repositories and model checkpoints, and was trained on the MIMIC-CXR dataset, not the Open-i IU X-ray dataset. \textbf{* indicates that the model was not conditioned on the report from the previous study, as the previous study is not available with the Open-i IU X-ray dataset.}}\label{tab:model_results_iu_x-ray}
\scriptsize

\begin{tabular}{lZ{14pt}Z{29pt}Z{29pt}Z{29pt}Z{29pt}Z{29pt}Z{29pt}Z{29pt}Z{29pt}Z{29pt}}

\toprule

\multirow{3}{*}{Model} & \multirow{2}{*}{Multi-} & \multicolumn{3}{c}{CheXbert} & \multirow{2}{*}{RadGraph} & \multirow{2}{*}{CXR-} & \multirow{3}{*}{BERTScore} & \multirow{3}{*}{CIDEr} & \multirow{3}{*}{ROUGE-L} & \multirow{3}{*}{BLEU-4} \\\cmidrule{3-5}
& image &F1 &P &R & ER & BERT & & \\
\midrule

R2Gen &\xmark &0.102 &\cellcolor[HTML]{fcecea}0.195 &0.101 &0.210 &0.544 &0.190 &0.078 &0.213 &\cellcolor[HTML]{fffcfc}0.023 \\
WCL &\xmark&\cellcolor[HTML]{d8f0e4}0.143 &0.161 &\cellcolor[HTML]{c3e7d5}0.167 &\cellcolor[HTML]{fffcfc}0.214 &\cellcolor[HTML]{f3faf7}0.555 &\cellcolor[HTML]{fffdfd}0.194 &0.077 &0.213 &0.022 \\
CvT2DistilGPT2 &\xmark&\cellcolor[HTML]{8ed2b0}0.220 &\cellcolor[HTML]{e67c73}\textbf{0.381} &\cellcolor[HTML]{97d5b7}0.214 &\cellcolor[HTML]{f8d7d5}0.258 &\cellcolor[HTML]{caeada}0.589 &\cellcolor[HTML]{f6cfcb}0.253 &\cellcolor[HTML]{c3e7d5}0.141 &\cellcolor[HTML]{f5cac6}0.255 &\cellcolor[HTML]{f6d0cd}0.039 \\
$\mathcal{M}^2$ ${\rm fact}_{\rm ENT}$ &\cmark &\cellcolor[HTML]{98d5b7}0.210 &\cellcolor[HTML]{f3bcb7}0.275 &\cellcolor[HTML]{68c296}0.265 &\cellcolor[HTML]{e67c73}\textbf{0.366} &\cellcolor[HTML]{eef9f4}0.558 &\cellcolor[HTML]{e78178}0.353 &\cellcolor[HTML]{57bb8a}\textbf{0.252} &\cellcolor[HTML]{e77f76}0.314 &\cellcolor[HTML]{e67c73}\textbf{0.068} \\
$\mathcal{M}^2$ ${\rm fact}_{\rm ENTNLI}$  &\cmark &\cellcolor[HTML]{81cca8}0.234 &\cellcolor[HTML]{ea8d85}0.354 &\cellcolor[HTML]{71c69c}0.256 &\cellcolor[HTML]{ec9890}0.334 &\cellcolor[HTML]{a4dabf}0.620 &\cellcolor[HTML]{e67c73}\textbf{0.359} &\cellcolor[HTML]{64c193}0.239 &\cellcolor[HTML]{e67c73}\textbf{0.316} &\cellcolor[HTML]{e88279}0.066 \\
CXRMate (ours)* &\cmark &\cellcolor[HTML]{57bb8a}\textbf{0.277} &\cellcolor[HTML]{ea8e87}0.351 &\cellcolor[HTML]{57bb8a}\textbf{0.283} &\cellcolor[HTML]{f2bbb7}0.291 &\cellcolor[HTML]{57bb8a}\textbf{0.683} &\cellcolor[HTML]{ec9992}0.323 &\cellcolor[HTML]{acdec6}0.164 &\cellcolor[HTML]{efa7a1}0.282 &\cellcolor[HTML]{f3bdb8}0.046 \\

\bottomrule
\end{tabular}

\end{table*}

\subsection{Comparison to Multi-Task Learning models}\label{sec:multi_task_results}

Here, we test against the multi-task learning models presented in Table \ref{tab:multi_task}, of which XrayGPT, LLM-CXR, and MedXChat leverage LLMs. We compare to the results from \citet{yang_medxchat_2023}, due to lack of code and model availability. First, in Table \ref{tab:multi_task} we highlight the discrepancies between each of the models evaluated in \citet{yang_medxchat_2023}. Each splits the MIMIC-CXR test set differently, with some excluding based on missing sections, and others excluding based on the view of the image. Moreover, there were differences between how the findings and impression sections are treated. Some only used one section, others combined the sections, and some differentiated between the sections. Each also formatted the radiologist reports differently, with some having unclear formatting rules. And finally, some were single-image models, and generated a report per image, while others were multi-image models, and generated a report per study.


\newcommand*{\gra}[1]{%
  \includegraphics[scale=0.3]{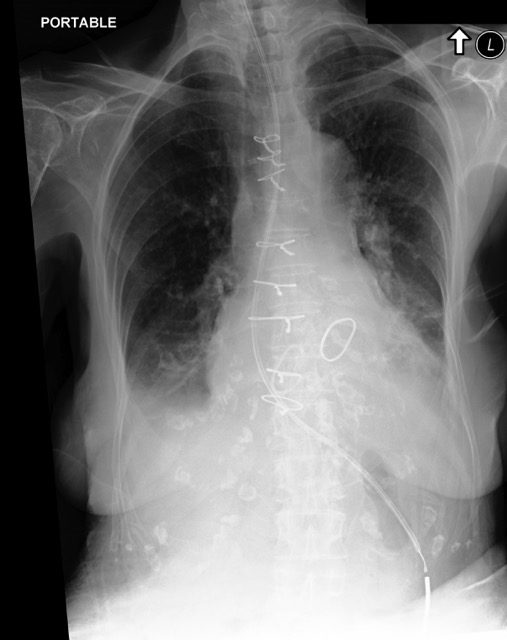}
}
\newcommand*{\grb}[1]{%
  \includegraphics[scale=0.3]{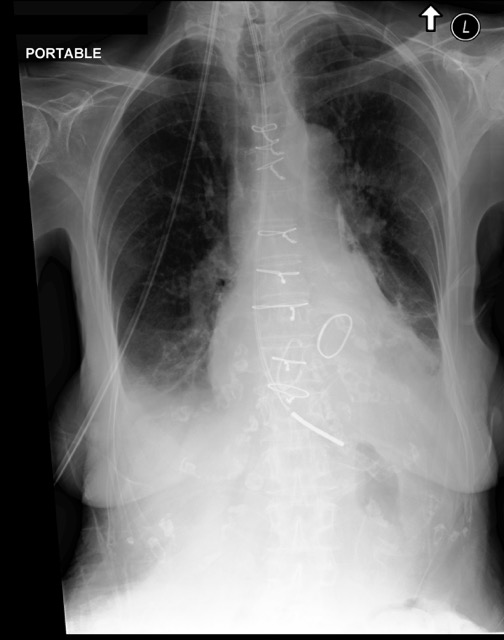}
}
\newcommand*{\grc}[1]{%
  \includegraphics[scale=0.3]{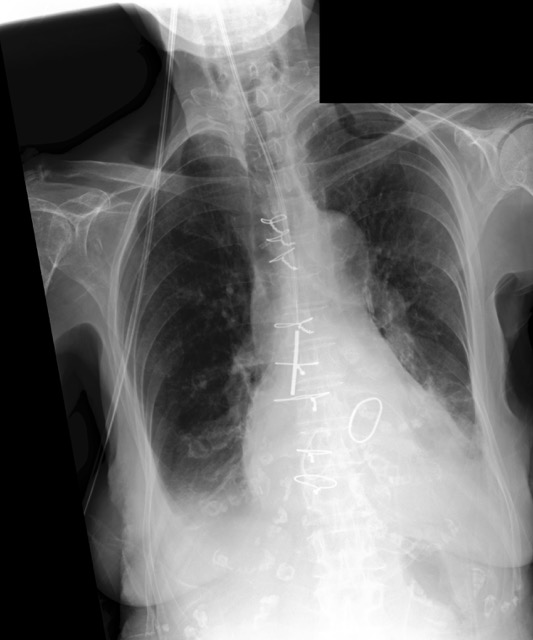}
}

\newcommand*{\horzcat}{\grc{c}\grb{b}\gra{a}}

\begin{table*}[]
\centering
\renewcommand{\arraystretch}{1.0}
\caption{\label{tab:case_study}A study where CXRMate attained a high CheXbert score. Shown are the three CXRs included with the study, along with a radiologist's report and generated reports. Matching highlighting indicates true positive observations between the generated and radiologist's report. The report generated by CXRMate was the most similar to the radiologist's report and only failed to detect the tricuspid heart valve replacement. The CheXbert score was calculated between the findings sections only.}

\begin{tabular}{p{6.6in}} 

\horzcat \\
\toprule

Radiologist's report (\textit{study }\texttt{59144799}\textit{ of patient }\texttt{18224196}\textit{. Their previous study (}\texttt{53536595}\textit{) was 59 hours prior}.)\\
\baselineskip=0.75\baselineskip

\textbf{Findings:}
\hla{Feeding tube tip in the distal stomach}. Central line, endotracheal tube have been removed. \hlb{Sternotomy}, \hlc{valve replacements}. \hld{Small bilateral pleural effusions have worsened}. \hle{Left basilar atelectasis or infiltrate, worsened}. \hle{Right basilar atelectasis, worsened}. \hlf{Increased heart size, more prominent}. \hlg{Mildly prominent pulmonary vascularity}.\\
\baselineskip=0.75\baselineskip

\textbf{Impression:}
\hla{Feeding tube tip in the distal stomach}. Worsened pulmonary findings\\

\hline

CXRMate (ours) \underline{\textit{CheXbert F1 score: 1.000}} \\
\baselineskip=0.75\baselineskip

\textbf{Findings:}
The patient is \hlb{status post median sternotomy} and \hlc{mitral valve replacement}. \hla{A Dobhoff tube terminates in the stomach}. There are \hld{small to moderate bilateral pleural effusions} with \hle{subjacent atelectasis}. The \hlf{heart is mildly enlarged}. There is \hlg{mild pulmonary vascular congestion}. No pneumothorax is identified. There are no acute osseous abnormalities.\\
\baselineskip=0.75\baselineskip

\textbf{Impression:}
\hla{The Dobhoff tube terminates in the stomach}. \hld{Small to moderate bilateral pleural effusions} with \hle{subjacent atelectasis}. \hlg{Mild pulmonary vascular congestion}.\\

\hline

\textbf{$\mathcal{M}^2$ ${\rm fact}_{\rm ENTNLI}$} \underline{\textit{CheXbert F1 score: 0.857}} \\
\baselineskip=0.75\baselineskip

\textbf{Findings:}
the cardiomediastinal and hilar contours are stable . \hla{the nasogastric tube ends in the stomach} . the lung volumes are lower compared to the prior chest radiograph . \hld{small bilateral pleural effusions are unchanged . small right pleural effusion is unchanged} . \hle{there is bibasilar atelectasis} . there is no pneumothorax . \hla{a dobbhoff tube is seen with the tip in the stomach} .\\

\hline

CvT2DistilGPT2 \underline{\textit{CheXbert F1 score: 0.333}} \\
\baselineskip=0.75\baselineskip

\textbf{Findings:}
as compared to the previous radiograph the patient has received a \hla{dobbhoff catheter. the course of the catheter is unremarkable the tip of the catheter projects over the middle parts of the stomach}. there is no evidence of complications notably no pneumothorax. otherwise the radiograph is unchanged.\\

\bottomrule

\end{tabular}
\end{table*}

Given these discrepancies, claims made from a comparison between these models will be weak. Nevertheless, we compare to these multi-task learners in Tables \ref{tab:medxchat_ce_results} and \ref{tab:medxchat_nlg_results}. Here, F + I indicates that the findings and impression sections were combined, while F \& I indicates that the sections could be differentiated after generation. CXRMate attained the highest micro and macro-averaged CheXpert F1 scores in Table \ref{tab:medxchat_ce_results}, as well as the highest scores for \textit{pneumothorax}, \textit{edema}, \textit{consolidation or pneumonia}, and \textit{lung lesion}. This weakly indicates that CXRMate is more diagnostically accurate than the multi-task learning models. Each model in Table \ref{tab:medxchat_nlg_results} attained the highest score for one metric, likely due to the discrepancies between the evaluation procedures of each model.

\newcommand*{\grd}[1]{%
  \includegraphics[scale=0.25]{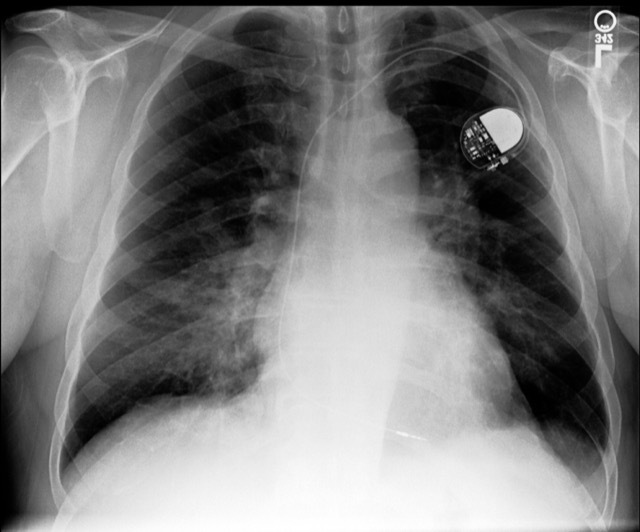}
}
\newcommand*{\gre}[1]{%
  \includegraphics[scale=0.25]{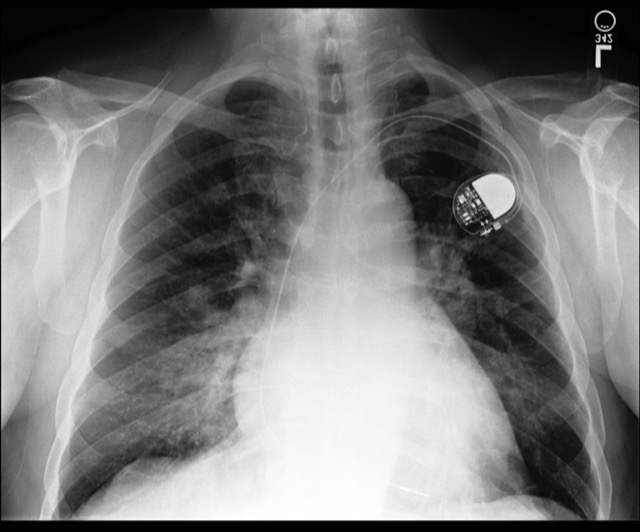}
}
\newcommand*{\grf}[1]{%
  \includegraphics[scale=0.25]{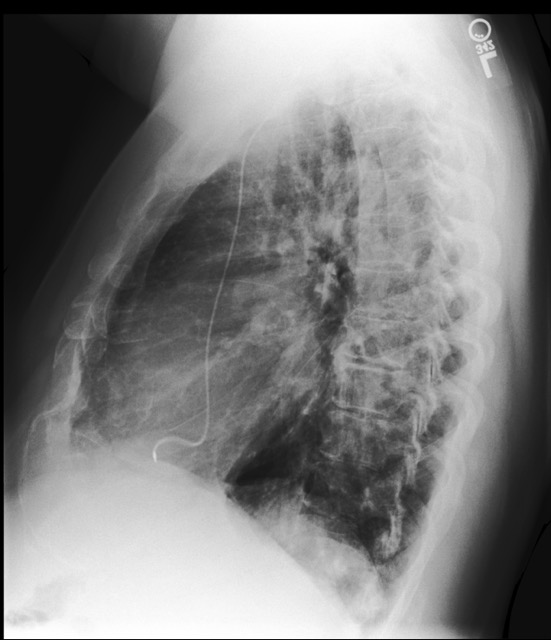}
}

\newcommand*{\horzcate}{\grd{d}\gre{e}\grf{f}}

\begin{table*}[]
\centering
\renewcommand{\arraystretch}{1.0}
\caption{\label{tab:case_study_2}A study where CXRMate attains a low CheXbert score. Shown are the three CXRs included with the study, along with a radiologist's report and generated reports. Matching highlighting indicates true positive observations between the generated and radiologist's report. The CheXbert score was calculated between the findings sections only.}

\begin{tabular}{p{6.6in}} 

\horzcate \\

\toprule

Radiologist's report (\textit{study }\texttt{59698565}\textit{ of patient }\texttt{18570152}\textit{. There was no prior study available}.)\\
\baselineskip=0.75\baselineskip

\textbf{Findings:}
Frontal and lateral views of the chest are obtained. \hla{Right lower lobe opacity is worrisome for consolidation, possibly due to pneumonia}. \hlb{Along the superior aspect of the right lower lung consolidation, there is a 0.9-cm nodular opacity, projecting between the posterior right sixth and seventh ribs, which could relate to consolidation or an underlying pulmonary nodule is not excluded}. Recommend followup chest radiograph after appropriate therapy and if finding remains, chest CT. There is also \hlc{a left suprahilar opacity, which could be a second site of infection or relate to mild volume overload}. \hld{There is central pulmonary vascular engorgement}. No large pleural effusion or pneumothorax is seen. \hle{Single-lead left-sided pacemaker is seen with leads in the expected position of the right ventricle}. \hlf{The cardiac silhouette is enlarged}.\\
\baselineskip=0.75\baselineskip

\textbf{Impression:}
\hla{Right lower lobe and left infrahilar opacities, right greater than left, in the appropriate clinical setting, raises concern for pneumonia}. Recommend followup to resolution. \hlb{Possible 0.9 cm nodular opacity along the superior aspect of the right lower lung opacity, could relate to consolidation, but pulmonary nodule not excluded}. Recommend followup chest radiographs after appropriate therapy and if finding remains, chest CT. \hlc{Left suprahilar opacity, which could be a second site of infection or relate to mild volume overload}. \hld{Pulmonary vascular engorgement}. \hlf{Enlarged cardiac silhouette}.\\
\hline

CXRMate (ours) \underline{\textit{CheXbert F1 score: 0.333}} \\
\baselineskip=0.75\baselineskip

\textbf{Findings:}
Frontal and lateral views of the chest were obtained. \hla{There is new right middle lobe consolidation worrisome for pneumonia}. The left lung is clear. There is no pleural effusion or pneumothorax. The cardiac and mediastinal silhouettes are stable. \hle{Single lead left-sided pacemaker is seen with lead extending to the expected position of the right ventricle}.\\
\baselineskip=0.75\baselineskip

\textbf{Impression:}
\hla{Right middle lobe consolidation worrisome for pneumonia}. Recommend followup to resolution.\\
\hline

$\mathcal{M}^2$ ${\rm fact}_{\rm ENTNLI}$ \underline{\textit{CheXbert F1 score: 0.333}} \\
\baselineskip=0.75\baselineskip

\textbf{Findings:}
the cardiomediastinal and hilar contours are normal . \hlf{the heart is mildly enlarged} . the lung volumes are lower compared to the prior chest radiograph . there is mild pulmonary edema . there is no focal consolidation . no pleural effusion or pneumothorax is seen . \hle{a left chest pacemaker is seen with leads in the} right atrium and \hle{right ventricle} .\\
\hline

CvT2DistilGPT2 \underline{\textit{CheXbert F1 score: 0.400}} \\
\baselineskip=0.75\baselineskip

\textbf{Findings:}
frontal and lateral views of the chest were obtained. \hle{single-lead left-sided pacemaker is again seen with lead extending to the expected position of the right ventricle}. \hlb{bilateral perihilar opacities right greater than left are worrisome for multifocal pneumonia}. no pleural effusion or pneumothorax is seen. cardiac and mediastinal silhouettes are stable.\\
\bottomrule

\end{tabular}
\end{table*}

\begin{table*}[]
\centering
\renewcommand{\arraystretch}{1.0}
\caption{\label{tab:label_issues}A study highlighting the issues introduced by the labels of \citet{chen_generating_2020}. Shown is a CXR from a study, along with a radiologist's report and generated reports. Matching highlighting indicates true positive observations between the generated and radiologist's report. The CheXbert score was calculated between the findings sections only.}

\begin{tabular}{c} 

\includegraphics[scale=0.4]{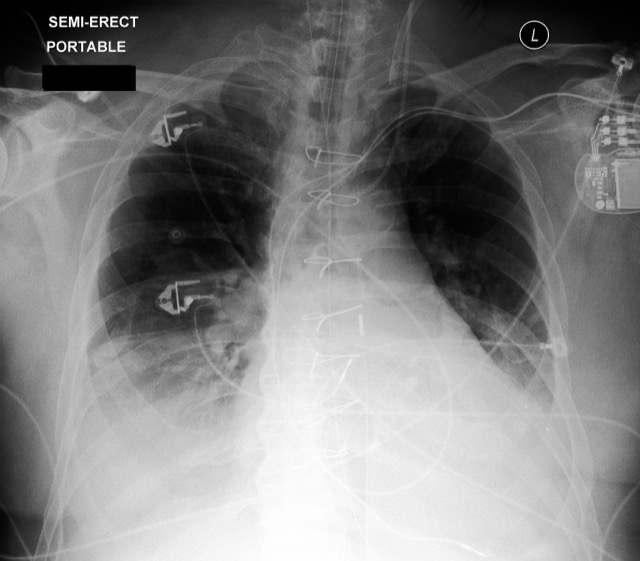}

\end{tabular}


\begin{tabular}{p{6.6in}} 

\toprule
Radiologist's report (\textit{study} \texttt{54849848} \textit{of patient} \texttt{10886362}. \textit{Their previous study (}\texttt{56034024}\textit{) was 3 days and 21 hours prior.})\\
\baselineskip=0.75\baselineskip

\textbf{Findings:}
AP single view of the chest has been obtained with patient in sitting semi-upright position. Comparison is made with the next preceding portable chest examination with the patient in supine position as of \_\_\_. Again noted is \hlh{status post sternotomy} and \hlg{significant enlargement of the cardiac silhouette}. Previously described \hlf{permanent pacer in left axillary position with two intracavitary electrodes in unchanged location}. \hle{Unchanged position of left internal jugular approach central venous line terminating in upper portion of SVC}. No pneumothorax has developed. \hld{Diffuse haze over both lung bases as before obliterating the diaphragmatic contours and indicative of bilateral pleural effusions partially layering posteriorly}. \hlc{The pulmonary venous congestive pattern persists}. \hlb{An intra-aortic balloon pump device is seen to terminate in the descending thoracic aorta about 3 cm below the level of the lower thoracic arch contour}. This is unchanged.\\
\baselineskip=0.75\baselineskip

\textbf{Impression:}
No significant interval changes during the last 24 hours interval. The described changes with \hlh{postoperative status}, \hla{CHF}, \hld{pleural effusion} and \hlb{intra-aortic balloon pump device in place} is of course compatible with the patient's hypoxia.\\
\hline

CXRMate (ours) \underline{\textit{CheXbert F1 score: 0.75}}\\
\baselineskip=0.75\baselineskip

\textbf{Findings:}
AP single view of the chest has been obtained with patient in semi-upright position. The patient is now intubated, the ETT terminating in the trachea 5 cm above the level of the carina. A right internal jugular approach central venous line is seen to terminate in the mid portion of the SVC. \hle{A left internal jugular approach central venous line remains in unchanged position terminating in the mid portion of the SVC}. An NG tube has been passed and reaches below the diaphragm, and terminates in the fundus of the stomach with side port remaining in the distal esophagus. \hlh{The patient is status post sternotomy}, CABG, and mitral valve prosthesis. \hlg{The heart is mildly enlarged} and the pulmonary vasculature is not congested. There are \hld{bilateral pleural effusions, slightly more so on the right than the left}. There is no pneumothorax and no new parenchymal opacities are present in the apical area.\\
\baselineskip=0.75\baselineskip

\textbf{Impression:}
NG tube reaches below the diaphragm, and side port is in the distal esophagus. \hlh{Post-CABG changes}. \hld{Bilateral pleural effusions, slightly larger on the right than the left} and mild on the right, and mild on the left, are stable.\\
\hline

CvT2DistilGPT2 \underline{\textit{CheXbert F1 score: 0.286}}\\
\baselineskip=0.75\baselineskip

\textbf{Findings:}
as compared to the previous radiograph the patient has been intubated. the tip of the endotracheal tube projects 4 cm above the carina. the patient has also received a nasogastric tube. the course of the tube is unremarkable the tip of the tube is not visualized on the image. there is no evidence of complications\\

\bottomrule

\end{tabular}
\end{table*}





\subsection{Generalisability to the Open-i IU X-ray dataset}

Here, we assess the generalisability of CXRMate, which has been trained on MIMIC-CXR, to another dataset, specifically, Open-i IU X-ray. Currently, the only publicly-available dataset where longitudinal data can be leveraged is MIMIC-CXR. Open-i IU X-ray only includes one study per patient, preventing longitudinal data from being leveraged. This is disadvantageous for CXRMate, as it cannot condition on the report from the previous study (see Table \ref{tab:history_impact}). Nevertheless, we evaluate the performance of CXRMate on Open-i IU X-ray, as shown in Table \ref{tab:model_results_iu_x-ray}. CXRMate generates both the findings and impression sections, while the remaining models in Table \ref{tab:model_results_iu_x-ray} generate only the findings section. Therefore, only the findings section for CXRMate was evaluated in Table \ref{tab:model_results_iu_x-ray} (against the findings section from the radiologist reports), and the impression section was ignored.

CXRMate produced the highest CheXbert (F1 and R) and CXR-BERT scores, indicating that it was able to generalise well in terms of clinical semantic similarity to the radiologist reports. However, $\mathcal{M}^2$ ${\rm fact}_{\rm ENTNLI}$ and $\mathcal{M}^2$ ${\rm fact}_{\rm ENT}$ attained a higher BERTScore, CIDEr, ROUGE-L, and BLEU-4, indicating that CXRMate was not able to generalise as well in terms of general semantic and syntactical similarity to the radiologist reports. This may have been exacerbated by the unavailability of the previous studies during generation. While $\mathcal{M}^2$ ${\rm fact}_{\rm ENTNLI}$ attained the highest BERTScore and ROUGE-L scores, $\mathcal{M}^2$ ${\rm fact}_{\rm ENT}$ attained the highest RadGraph ER, CIDEr, and BLEU-4 scores. Previously, in Table \ref{tab:model_results}, $\mathcal{M}^2$ ${\rm fact}_{\rm ENTNLI}$ outperformed $\mathcal{M}^2$ ${\rm fact}_{\rm ENT}$ on each metric with the MIMIC-CXR test set. This indicates that $\mathcal{M}^2$ ${\rm fact}_{\rm ENT}$ was able to better generalise from the MIMIC-CXR test set to the Open-i IU X-ray dataset than $\mathcal{M}^2$ ${\rm fact}_{\rm ENTNLI}$.

\subsection{Case Studies}

In Table \ref{tab:case_study}, we show a study where CXRMate attained a high CheXbert F1 score. The study is of a feeding tube being inserted into the stomach. It includes three CXRs and is an example of why it is important to condition on all images of a study, even if they are all the same view. Along with the feeding tube, this study includes postoperative hardware, mitral and tricuspid heart valve replacements, and several pathologies. CXRMate successfully identified most of these (minus the tricuspid heart valve replacement). $\mathcal{M}^2$ ${\rm fact}_{\rm ENTNLI}$ was only able to identify the feeding tube, the small bilateral pleural effusions, and the bibasilar atelectasis. CvT2DistilGPT2 was only able to identify the feeding tube. At least for this study, CXRMate was able to generate a succinct, intelligible report that was factually more correct than other models with respect to the radiologist's report.

In Table \ref{tab:case_study_2}, we show a study where CXRMate attained a low CheXbert F1 score. This study includes several findings identified by the radiologist. While each model was able to identify the pacemaker, its leads, and their location, they all struggled to identify the remaining findings. This example contains multiple CheXbert observations that CXRMate (as well as the other models) performed poorly on in Figure \ref{fig:pathology_results}, including \textit{lung opacity} (CheXbert F1 of 0.491), \textit{pneumonia} (CheXbert F1 of 0.235), \textit{consolidation} (CheXbert F1 of 0.194), and \textit{lung lesion} (or nodule) (CheXbert F1 of 0.077). CXRMate only identified the consolidation and potential pneumonia; however, the location of the consolidation was inconsistent with the radiologists determination (the radiologist determined that the consolidation was in the right lower lung, whereas CXRMate determined that it was in the right middle lobe). $\mathcal{M}^2$ ${\rm fact}_{\rm ENTNLI}$ and CvT2DistilGPT2 also failed to identify several of the radiologist's findings. This demonstrates some of the limitations of current CXR report generation models.

In Table \ref{tab:label_issues}, we show a study that highlights the issues caused by the labels of \citet{chen_generating_2020}. As previously highlighted in Tables \ref{tab:missing_chen_labels} and \ref{tab:trunc_chen_labels}, the labels of \citet{chen_generating_2020} do not include an impression section and as a result have information loss compared to the original findings sections when there are more than 100 tokens. To show the impact of this on a CXR report generator, we selected a study with a longer radiologist report, and compared it to the report generated by CvT2DistilGPT2, which was the best performing model trained on the labels of \citet{chen_generating_2020}. First, the impression section was not generated, which is a standard section of a radiologist report. Second, CvT2DistilGPT2 struggled with generating reports of longer length, and was unable to capture any of the findings mentioned in the radiologist's report. Contrary to this, CXRMate, which was trained on both the findings and impression sections with minimal formatting, was able to generate a longer findings section and capture multiple findings that were mentioned in the radiologist's report. It also generated the impression section.

\section{Limitations and Future Directions}\label{sec:future}

\begin{figure}[]
    \centering
    
        \includegraphics[scale=0.85]{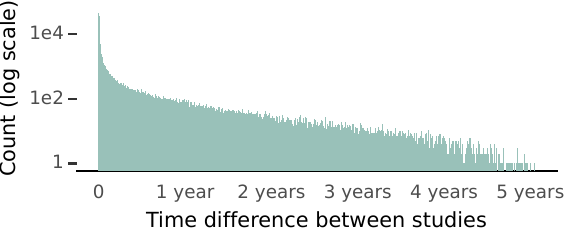}
        
    \caption{\label{fig:time_delta}Histogram of the training split of MIMIC-CXR of time difference between patients' studies.}
    
\end{figure}

Due to the difficulty in obtaining publicly-available datasets of quality that also retain longitudinal information between studies, our evaluation was limited to MIMIC-CXR and Open-i IU X-ray, the later of which does not contain longitudinal information. We aim to source additional datasets in future studies of high quality, possibly from a private collection. As with others in the literature, we use the JPG and PNG versions of MIMIC-CXR and Open-i IU X-ray, respectively, and resize the images to a lower resolution, which deviates from the quality of the images that a radiologist would be interpreting. Using the DICOM versions of these datasets would reduce quantisation error, while using a higher resolution could reduce the risk of removing fine details. We hypothesise that these factors could be beneficial for CXR report generation, and should be considered in future work. While our metrics are correlated with radiologists’ assessment of reporting, we aim to source practising radiologists for qualitative evaluation of the generated reports in future investigations. In preliminary testing, we provided the time difference between the current and previous study to the model, however, this had no impact on performance. The time differences between studies from MIMIC-CXR are shown in Figure \ref{fig:time_delta}; there can be a large difference in time between studies, which may impact its efficacy as a feature. We also did not consider the images from previous studies, or a history size larger than just the previous study. We aim to explore these in future investigations. The CXR-BERT reward in its current form can lead to repetitions in the generated report for some training runs. We aim to mitigate this by adding a brevity penalty.

\section{Conclusion} 

We demonstrate that our proposed model, CXRMate, generates radiology reports that are more closely aligned with those of radiologists than current state-of-the-art models, such as those leveraging LLMs, reinforcement learning, and multi-task learning. We also demonstrate that conditioning on longitudinal data when available, and on all images of a study, improves CXR report generation. Moreover, we show that the CXR-BERT reward is a promising alternative to the state-of-the-art RadGraph ER reward. We also demonstrate that differentiating each section with section embeddings improves CXR report generation. Furthermore, we highlight issues pertaining to the evaluation of a large portion of CXR report generators in the literature, caused by excessive formatting. Finally, we open source CXRMate to encourage reproducibility. By improving diagnostic accuracy, we hope that CXRMate brings CXR report generation a step closer to clinical trial consideration.

\bibliographystyle{model2-names}
\bibliography{references}

\end{document}